\documentclass{article}
\usepackage[ruled,lined,norelsize,linesnumbered]{algorithm2e}

\usepackage{microtype}
\usepackage{graphicx}
\usepackage{booktabs}  
\usepackage{hyperref}

\usepackage[accepted]{icml2023}
\usepackage{amsmath}
\usepackage{amssymb}
\usepackage{mathtools}
\usepackage{amsthm}

\theoremstyle{plain}
\newtheorem{theorem}{Theorem}

\theoremstyle{definition}
\newtheorem{definition}[theorem]{Definition}

\theoremstyle{remark}

\icmltitlerunning{Pareto Manifold Learning: Tackling multiple tasks via ensembles of single-task models}

\newtheorem{claim}[theorem]{Claim}
\usepackage{_abbreviations_}
\usepackage{_mathcmds_}
\usepackage{_cmds_}
\usepackage{paralist}
\usepackage{etoolbox}
\usepackage{betaplot}
\usepackage{siunitx}
\usepackage{enumitem}
\usepackage{caption}
\usepackage{subcaption}
\usepackage{float}

\makeatletter
\patchcmd{\@algocf@start}{-1.5em}{0pt}{}{}
\newcommand\Autoref[1]{\@first@ref#1,@}
\def\@throw@dot#1.#2@{#1}%
\def\@set@refname#1{%
    \edef\@tmp{\getrefbykeydefault{#1}{anchor}{}}%
    \xdef\@tmp{\expandafter\@throw@dot\@tmp.@}%
    \ltx@IfUndefined{\@tmp autorefnameplural}%
         {\def\@refname{\@nameuse{\@tmp autorefname}s}}%
         {\def\@refname{\@nameuse{\@tmp autorefnameplural}}}%
}
\def\@first@ref#1,#2{%
  \ifx#2@\autoref{#1}\let\@nextref\@gobble%
  \else%
    \@set@refname{#1}%
    \@refname~\ref{#1}%
    \let\@nextref\@next@ref%
  \fi%
  \@nextref#2%
}
\def\@next@ref#1,#2{%
   \ifx#2@ and~\ref{#1}\let\@nextref\@gobble%
   \else, \ref{#1}%
   \fi%
   \@nextref#2%
}
\makeatother

\newcommand{\conditionalclearpage}{\clearpage}
\newcommand{\myparagraph}[1]{\paragraph{#1}}
\newcommand{\abbrevmtl}{\shortmtl}

\SetCommentSty{mycommfont}
\SetKwInOut{Input}{Input}
\SetKwInOut{Output}{O}

\SetAlgoLined
\SetAlCapHSkip{0pt} 

\begin{document}

\twocolumn[
\icmltitle{Pareto Manifold Learning: Tackling multiple tasks via ensembles of single-task models}

\icmlsetsymbol{equal}{*}

\begin{icmlauthorlist}
\icmlauthor{Nikolaos Dimitriadis}{epfl}
\icmlauthor{Pascal Frossard}{epfl}
\icmlauthor{Fran\c{c}ois Fleuret}{unige}
\end{icmlauthorlist}

\icmlaffiliation{epfl}{Ecole Polytechnique F\'ed\'erale de Lausanne (EPFL), Lausanne, Switzerland}
\icmlaffiliation{unige}{University of Geneva, Geneva, Switzerland}

\icmlcorrespondingauthor{Nikolaos Dimitriadis}{nikolaos.dimitriadis@epfl.ch}

\icmlkeywords{Machine Learning, ICML}

\vskip 0.3in
]

\printAffiliationsAndNotice{}  %

\newif\ifshowappendix
\showappendixfalse

\begin{abstract}
In Multi-Task Learning (MTL), tasks may compete and limit the performance achieved on each other, rather than guiding the optimization to a solution, superior to all its single-task trained counterparts.
Since there is often not a unique solution optimal for all tasks, practitioners have to balance tradeoffs between tasks' performance, and resort to optimality in the Pareto sense.
Most MTL methodologies either completely neglect this aspect, and instead of aiming at learning a Pareto Front, produce one solution predefined by their optimization schemes, or produce diverse but discrete solutions.
Recent approaches parameterize the Pareto Front via neural networks, leading to complex mappings from tradeoff to objective space.
In this paper, we conjecture that the Pareto Front admits a linear parameterization in parameter space, which leads us to propose \textit{\pmlfull}, an ensembling method in weight space.
Our approach produces a continuous Pareto Front in a single training run, that allows to modulate the performance on each task during inference.
Experiments on multi-task learning benchmarks, ranging from image classification to tabular datasets and scene understanding, show that \textit{\pmlfull} outperforms state-of-the-art single-point algorithms, while learning a better Pareto parameterization than multi-point baselines.
\end{abstract}

\section{Introduction} 
\label{sec:intro}
In \mtl (\shortmtl), multiple tasks are learned concurrently within a single model,  striving towards infusing inductive bias that will help outperform the single-task baselines.
Apart from the promise of superior performance and some theoretical benefits \citep{Ruder_2017a}, such as generalization properties for the learned representation, modeling multiple tasks jointly has practical benefits as well, e.g., lower training and inference times and memory requirements.
However, building machine learning models presents a multifaceted host of decisions for multiple and often competing objectives, such as model complexity, runtime and generalization. Conflicts arise since optimizing for one metric often leads to the deterioration of other(s). 
A single solution satisfying optimally all objectives rarely exists and practitioners must balance the inherent trade-offs. 

The notion of tradeoffs is formally defined as \textit{Pareto optimality}.
In contrary to single-task learning, where one metric governs the comparison between methods (e.g., top-1 accuracy in \texttt{ImageNet}), multiple models can be optimal  in \abbrevmtl; e.g., model X yields superior performance on task \( \calA \) compared to model Y, but the reverse holds true for task \( \calB \); thus, there is not a single better model among the two.  Intuitively, improvement on an individual task performance can come only at the expense of another task.
\looseness=-1

\looseness=-1

In this paper, we develop a novel method, \textit{\pmlfull}, which casts \abbrevmtl problems as learning an ensemble of single-task predictors by interpolating among (ensemble) members during training. 
By operating in the convex hull of the members' weight space, each single-task model infuses and benefits from representational knowledge to and from the other members. During training, the losses are weighted in tandem with the interpolation, i.e., a monotonic relationship is imposed between the degree of a single-task predictor participation and the weight of the corresponding task loss. Consequently, the ensemble as a whole engenders a (weight) subspace that explicitly encodes tradeoffs and results in a continuous parameterization of the Pareto Front. We identify challenges in guiding the ensemble to such subspaces, designated \textit{Pareto subspaces}, and propose solutions regarding balancing the loss contributions, and regularizing the Pareto properties of the subspaces and adapting the interpolation sampling distribution.
\looseness=-1

Our method is based on a novel geometrical perspective; multiple Pareto stationary points lie in close proximity and are connected by simple paths whose parameterization produces a monotonic mapping in objective space.
This is motivated by the recent advancements in single task machine learning that have explored the geometry of the loss landscape and shown experimentally that local optima are connected by simple paths, even linear ones in some cases \citep{Wortsman_Horton_Guestrin_etal_2021,Garipov_Izmailov_Podoprikhin_etal_2018,Frankle_Dziugaite_Roy_etal_2020,Draxler_Veschgini_Salmhofer_etal_2018}. We assume that, when the problem has multiple objectives, it acquires a new dimension relating to the number of tasks. Concretely, there are multiple loss landscapes and a solution that satisfies users' performance requirements must lie in the intersection of low loss valleys (for all tasks). 
\looseness=-1

\looseness=-1

Experimental results validate that the proposed method is able to discover \textit{Pareto Subspaces}, and outperforms baselines on multiple benchmarks. Our training scheme offers two main advantages. First, enforcing low loss for all tasks on a linear subspace implicitly penalizes curvature, which has been linked to generalization \citep{Chaudhari_Choromanska_Soatto_etal_2017}, benefitting all tasks' performance. Second, the algorithm produces in a single training run and with minimal additional complexity a subspace of Pareto Optimal solutions, rather than a single model, enabling practitioners to handpick during inference the solution that offers the tradeoff that best suits their needs. The source code is available at \url{https://github.com/nik-dim/pamal}.

Overall, our main contributions are the following:
\begin{itemize}[topsep=0pt]
    \setlength\itemsep{-1pt}
    \item we offer a geometrical view on the problem of Pareto Front Learning and show that multiple functionally diverse solutions can exist in a straight path in weight space (\autoref{sec:problem formulation}), 
    \item we propose a novel algorithm, \pmlfull, that employs weight ensembles to infuse inductive bias to the optimization trajectory regarding the monotonicity dictated by Pareto optimality (\autoref{sec:method}),
    \item We validate our approach on several benchmarks and show that it outperforms baselines, while producing a more reliable mapping from desired preference to objective space compared to other Pareto Front Approximation techniques (\autoref{sec:experiments}).
\end{itemize}
\looseness=-1

\begin{figure*}[!t]
    \centering
    \includegraphics[width=\textwidth]{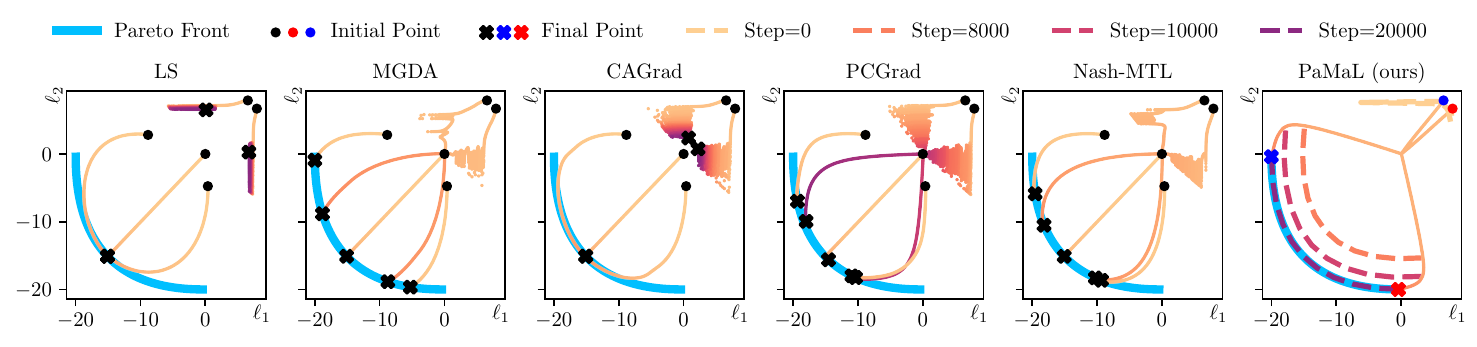}
    \protect\caption{Illustrative example following \cite{Yu_Kumar_Gupta_etal_2020,Navon_Shamsian_Achituve_etal_2022}. We present the optimization trajectories in loss space starting from different initializations (black bullets) leading to final points (crosses). Color reflects the iteration number when the corresponding value is achieved.  
    To highlight that our method (\pml) deals in pairs of models, we use blue and red to differentiate them. Dashed lines show intermediate results of the discovered subspace. While baselines may not reach the Pareto Front or display bias towards specific solutions, \pml discovers the entire Pareto Front \textit{in a single run}  and shows superior functional diversity. 
    }
    \label{fig:illustrative}
\end{figure*}
\section{Related Work} 
\label{sec:related work}

\myparagraph{\mtl} Learning multiple tasks in the Deep Learning setting \cite{Ruder_2017a,Crawshaw_2020} is usually approached by architectural methodologies \citep{Misra_Shrivastava_Gupta_etal_2016,Ruder_Bingel_Augenstein_etal_2019}, where the architectural modules are combined in several layers to govern the joint representation learning,  or optimization approaches \citep{Cipolla_Gal_Kendall_2018,Chen_Badrinarayanan_Lee_etal_2018}, where the architecture is standardized to be an encoder-decoder(s), for learning the joint and task-specific representations, respectively, and the focus shifts to the descent direction for the shared parameters. We focus on the more general track of optimization methodologies fixing the architectural structure to Shared-Bottom \cite{Caruana_1997}. The various approaches focus on finding a suitable descent direction for the shared parameters. The optimization methods can be broadly categorized into \textit{loss-} and \textit{gradient-balancing}  \citep{Liu_Li_Kuang_etal_2020}. For the former, the goal is to appropriately weigh the losses, e.g.,  via task-dependent homoscedastic uncertainty \cite{Cipolla_Gal_Kendall_2018},  by enforcing task gradient magnitudes to have close norms \cite{Chen_Badrinarayanan_Lee_etal_2018}. 
The latter class of methodologies manipulate the gradients so that they satisfy certain conditions; projecting the gradient of a (random) task on the normal plane of another so that gradient conflict is avoided \cite{Yu_Kumar_Gupta_etal_2020},  enforcing the common descent direction to have equal projections for all task gradients \cite{Liu_Li_Kuang_etal_2020}, casting the gradient combination as a bargaining game \cite{Navon_Shamsian_Achituve_etal_2022}. 
While the aforementioned methodologies focus on the \textit{Single Input-Multiple Outputs} (SIMO) setting, \mtl can also be studied under the Multiple Input-Multiple Output prism \citep{long2017learning,shen2021variational}. In this case, the challenge lies in the dearth of training data and the goal also includes the characterization of task relatedness.

\myparagraph{\mtl for Pareto Optimality} 
\citet{Sener_Koltun_2018} were the first to view the search for a common descent direction under the Pareto optimality prism and employ the Multiple Gradient Descent Algorithm (MGDA) \cite{Desideri_2012} in the Deep Learning context. However, MGDA does not account for task preferences 
\cite{Lin_Zhen_Li_etal_2019}, and biases solutions towards the task with the smallest gradient magnitude \citep{Liu_Li_Kuang_etal_2020}. By solving a slightly different formulation, \citet{Lin_Zhen_Li_etal_2019} are able to systematically introduce task trade-offs and produce a \textit{discrete} Pareto Front. 
However, each point requires a different training run.
\citet{Ma_Du_Matusik_2020} propose an orthogonal approach for Pareto stationary points; after a model is fitted with any \abbrevmtl method, a separate phase seeks other Pareto stationary points in its vicinity. But training still needs to occur for every seed point, the separate phase overhead grows linearly with the number of additional models, and the Pareto Front is not continuous  across seed points in \textit{parameter space}.
\citet{Navon_Shamsian_Fetaya_etal_2021} and \citet{Lin_Yang_Zhang_Kwong_2021} employ hypernetworks to continuously approximate the Pareto Front in a single run, which introduces additional design choices
and suffers from limited scalability, due to the hypernetwork requiring multiple times the number of parameters of the target network. \citet{Ruchte_Grabocka_2021} address the scalability issues by augmenting the feature space 
with the desired trade-off, which sacrifices either functional diversity or optimality. In both cases, the connection between desired tradeoff and network weights is obfuscated by the complex dynamics of a forward pass by a full neural network, and may not comply to the monotonicity constraints of Pareto optimal sets of solutions. For a fixed training budget, it may be more beneficial to ignore the user preference and search for one weight configuration to dominate all. Our approach is based on weight ensembles which can be seen as a particular case of linear hypernetworks; instead of generating weights, a convex combination of stored parameters is performed. This change of perspective infuses geometrical inductive bias and allows for more reliable and scalable Pareto Front Learning.
\looseness=-1

\myparagraph{Ensemble Learning and Mode Connectivity}
\label{sec:mode connectivity}
Apart from \abbrevmtl, our algorithm is methodologically tied to prior work in the geometry of the single-task neural network optimization landscapes.
The authors in \citep{Garipov_Izmailov_Podoprikhin_etal_2018,Draxler_Veschgini_Salmhofer_etal_2018} independently and concurrently showed that for two local optima \( \thopt_1,\thopt_2 \) produced by separate training runs (but same initializations) there exist nonlinear paths, defined as \textit{connectors} by \citet{Wortsman_Horton_Guestrin_etal_2021}, where the loss remains low. The connectivity paths can be extended to include linear in the case of the training runs sharing some part of the optimization trajectory \cite{Frankle_Dziugaite_Roy_etal_2020}. These findings can be leveraged to train a neural network subspace by enforcing linear connectivity among the subspace endpoints \cite{Wortsman_Horton_Guestrin_etal_2021}.
Linear mode connectivity encourages flatness and, therefore, is linked with methods explicitly enforcing flat minima \citep{Chaudhari_Choromanska_Soatto_etal_2017,Foret_Kleiner_Mobahi_etal_2021,Dinh_Pascanu_Bengio_Bengio_2017,Jiang*_Neyshabur*_Mobahi_Krishnan_Bengio_2020}. These approaches are applicable when designing a single objective, e.g. average of losses in \mtl, but do not allow for the infusion of Pareto properties and the inclusion of tradeoffs.
\citet{Izmailov_Podoprikhin_Garipov_etal_2018} produce flat minima by averaging multiple weight vectors discovered during the optimization trajectory, so that the final model lies in the middle of the low-loss basin. \citet{Wortsman_Ilharco_Gadre_Roelofs_Lopes_Morcos_Namkoong_Farhadi_Carmon_Kornblith_et_al_2022} perform weight ensembling with fine-tuned models produced via different hyperparameter configurations.
Apart from the recent weight ensembling works, output ensembling has been one of the staples of machine learning literature.
\citet{Lakshminarayanan_Pritzel_Blundell_2017a} utilize deep ensembles for uncertainty prediction but inference scales linearly with the number of ensemble members.
\citet{Wen_Tran_Ba_2020} improve on the computational complexity of output ensembles by sharing the bulk of the parameters among members and differentiating them via rank-1 matrices, while \citet{Havasi_Jenatton_Fort_Liu_Snoek_Lakshminarayanan_Dai_Tran_2021} employ a multi-input multi-output network by accommodating independent subnetworks for each ensemble and allowing a single-forward pass ensemble prediction. 
However, this results in subnetworks with incompatible architecture which does not allow for a continuous approximation of the Pareto Front.
\looseness=-1
\section{Problem Formulation}
\label{sec:problem formulation}
\myparagraph{Notation}
\label{sec:notation}
We use bold font for vectors \( \xx \), capital bold for matrices \( \bm{X} \) and regular font for scalars \( x \). \numtasks is the number of tasks and \nummembers is the number of ensemble members. Each task \( t\in[\numtasks] \) has a loss \( \calL_t \). The overall multi-task loss is \( \vectorloss=\begin{bmatrix}
    \calL_1,\dots,\calL_\numtasks
\end{bmatrix}^\top  \). \( \ww \in \Delta_\numtasks \subset\bbR^{\numtasks} \) is the weighting scheme for the tasks, i.e., the overall loss is calculated as \( \calL=\ww ^\top \vectorloss =\sum_{t=1}^{\numtasks} w_t\calL_t  \). 
Each member \( k\in[\nummembers] \) is associated with parameters \( \nnparams{k}\in\r^N \) and weighting \( \ww\in\Delta_\numtasks \). 
 \looseness=-1

\myparagraph{Preliminaries}
\label{sec:preliminaries}
\looseness=-1
Our goal lies in solving an unconstrained vector optimization problem of minimizing \( \bm{L}(\yy, \hat{\yy})=[\loss_1(y_1, \hat{y}_1), \dots, \loss_\numtasks(y_\numtasks, \hat{y}_\numtasks)]^\top  \), where \( \loss_i \) corresponds to the objective function for the \( i^{\text{th}} \) task, e.g., cross-entropy loss in case of classification. Constructing an optimal solution for all tasks is often unattainable due to competing objectives. Hence, an alternative notion of optimality is used, as described in \autoref{def:pareto optimiality}.
\looseness=-1

\begin{definition}[Pareto Optimality]
    \label{def:pareto optimiality}
    Consider two points \( \xx \) and \( \yy \) in the parameter space.
    A point \( \xx \) dominates a point \( \yy  \)  if \( \LL_t\left(\xx \right) \leq \LL_t\left(\yy \right) \) for all tasks \( t \in[\numtasks]\) and \( \bm{L}\left(\xx\right) \neq \bm{L}\left(\yy \right) \). Then, a point \( \xx\) is called Pareto optimal if there exists no point \( \yy \) that dominates it.
    The set of Pareto optimal points forms the Pareto front \( \mathcal{ P }_{\bm{L}}  \).
\end{definition}

The vector loss function is scalarized by the vector \( \ww\in[0,1]^\numtasks \) to form the overall objective \( \ww ^\top \bm{L} \). Without loss of generality, we assume that \( \ww \) lies in the \( \numtasks \)-dimensional simplex \( \Delta_\numtasks \) by imposing the constraint \( \| \ww \|=\sum_{t=1}^{\numtasks} w_t=1  \). This formulation permits to think of the vector of weights as an encoding of task preferences, e.g., for two tasks letting \( \ww=[0.8, 0.2] \) results in attaching more importance to the first task. Overall, the \abbrevmtl problem can be formulated within the Empirical Risk Minimization (ERM) framework for preference vector \( \ww \) and dataset \( \calD = \{(\xx, \yy)\}_{i=1} \) as:
\looseness=-1
\begin{equation}
    \label{eq:erm-mtl}
    \min_{\qq}\quad \bbE_{(\xx,\yy)\sim \calD} \left[\bm{L}\left(\ff\left(\xx;\qq\right) , \yy\right)\right]
\end{equation}
Our overall goal is to discover a low-dimensional parameterization in weight space that yields a (continuous) Pareto Front in functional space. This desideratum leads us to the following definition:
\begin{definition}[Pareto Subspace]
    \label{def:pareto subspace}
    Let \( \numtasks \) be the number of tasks, \( \calX \) the input space, \( \calY \) the multi-task output space, \( \calR \subset \r^N\) the parameter space, \( f:\calX\times\calR\rightarrow \calY \) the function implemented by a neural network, and \( \bm{L}:\calY\times\calY\rightarrow \r_{>0}^\numtasks \) be the vector loss. Let \( \{\bm{\theta}_t \in\calR: t\in [T]\}\) be a collection of network parameters and \( \calS \) the corresponding convex envelope, i.e., \( \calS=\left\{\sum_{t=1}^{T} \alpha_t\bm{\theta}_t  :\sum_{t=1}^{T} \alpha_t=1 \textrm{ and } \alpha_t \geq 0,  \forall t \right\} \). For the dataset \( \calD=(\calD_\calX,\calD_\calY) \), the subspace \( \calS \) is called Pareto if its mapping to functional space via the network architecture \( f \) forms a Pareto Front \( \calP=\bm{L}(f(\calD_\calX; \calS),\calD_\calY)=\{\bm{l}: \bm{l}=\bm{L}(f(\calD_\calX; \bm{\theta}), \calD_\calY), \quad\forall \bm{\theta}\in \calS\}  \).
\end{definition}
\looseness=-1
\begin{figure}[t!]
    \centering
    \includegraphics[width=\linewidth]{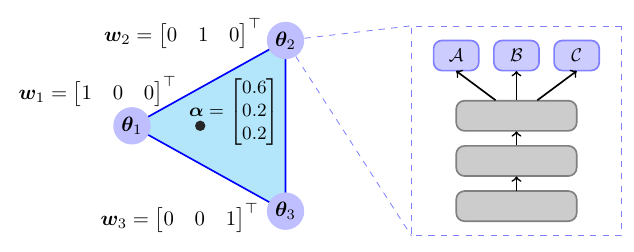}
    \noindent
    \caption{
        A representation of parameter space for \( \numtasks=3 \) tasks. Each node corresponds to a tuple of parameters and weighting scheme \( (\nnparams{v},\weighting{v})\in\r^N\times \Delta_\numtasks \). The blue dashed frame shows the model, e.g., shared-bottom architecture, implemented by the parameters \( \nnparams{v} \) of each node. For each training step, we sample \( \aa\in\Delta_\numtasks\) and construct the weight combination \( \bm{\theta}=\aa ^\top \bm{\Theta}=0.6\cdot\nnparams{1}+0.2\cdot\nnparams{2}+0.2\cdot\nnparams{3} \). 
    }
    \label{fig:simplex-graph-ll}
\end{figure}

\section{Method}
\label{sec:method}
We seek to find a collection of \nummembers neural networks, of identical architecture, whose linear combination in \textit{weight space} forms a continuous Pareto Front in \textit{objective space}. Model \( i \) corresponds to a tuple of network parameters \( \nnparams{i} \) and task weighting \( \weighting{i} \) and implements the function \( \ff(\cdot ; \nnparams{i}) \). 
We impose connectivity among models by modeling the subspace in the convex hull of the ensemble members.  \Autoref{sec:pml} presents the core of the algorithm, and in \Autoref{sec:algo extensions} we discuss various improvements that address \abbrevmtl challenges.
\looseness=-1

\subsection{Pareto Manifold Learning}
\label{sec:pml}

Let \( \paramscapital=\begin{bmatrix}
    \nnparams{1}, \nnparams{2}, \dots , \nnparams{\nummembers}
\end{bmatrix}^\top  \) be an \( \nummembers\times N \) matrix storing the parameters of all models, \( \bm{W}=\begin{bmatrix}\weighting{1}, \dots ,\weighting{\nummembers}\end{bmatrix} ^\top \) be a \( \nummembers\times\numtasks \) matrix storing the task weighting of ensemble members. By designing the subspace as a simplex, the objective now becomes:
\begin{equation}
    \label{eq:erm-simplex}
    \bbE_{(\xx,\yy)\sim \calD} \left[\bbE_{\aa\sim\textrm{P}}\left[ \aa ^\top \bm{W} \bm{L}\left(\ff\left(\xx;\aa\paramscapital\right), \yy \right)\right]\right]
\end{equation}
where \(\textrm{P}\) is the sampling distribution placed upon the simplex. In the case where the ensemble members are single-task predictors (\( \ww \) is one-hot) and the number of tasks coincides with the number of ensemble members \( (\nummembers=\numtasks) \), the matrix of task weightings \( \WW \) is an identity matrix and \autoref{eq:erm-simplex} simplifies to \(\bbE_{(\xx,\yy)\sim \calD} \left[\bbE_{\aa\sim\textrm{P}}\left[ \aa ^\top \bm{L}\left(\ff\left(\xx;\aa\paramscapital\right) , \yy\right)\right]\right] = \bbE_{(\xx,\yy)\sim \calD} \left[\bbE_{\aa\sim\textrm{P}}\left[ \sum_{t=1}^{T} \alpha_t\calL_t \left(\ff\left(\xx;\sum_{t=1}^T \alpha_t\nnparams{t}\right), \yy\right)\right]\right]\). By using the same weighting for both the losses and the ensemble interpolation, we explicitly associate models and task losses with a one-to-one correspondence, infusing preference towards one task rather than the other and guiding the learning trajectory to a subspace that encodes such tradeoffs.

\Autoref{algo:full} presents the full training procedure for this ensemble of neural networks, containing modifications discussed in subsequent sections. \Autoref{fig:illustrative} showcases the algorithm in a toy example. Consider an ensemble parameterized by \( \bm{\Theta}=\begin{bmatrix}
    \nnparams{1} &\cdots&\nnparams{\numtasks}
\end{bmatrix}^\top\). Concretely, at each training step with inputs \( \mathbf{x} \) and targets \( \mathbf{y} \) a random \( \aa \) is sampled and the corresponding convex combination of the networks is constructed \( \bmth = \aa ^\top  \bm{\Theta} \) (\Autoref{line:construct convex hull}). This procedure is shown in \Autoref{fig:simplex-graph-ll}. The batch is forwarded through the constructed network and the vector loss is scalarized by \( \aa \) as well, as in \Autoref{line:calculate loss}. The procedure is repeated \( W \) times at each batch (see \Autoref{sec:algo extensions}) and a regularization term penalizing non-Pareto stationary points is added (\Autoref{line:reg multiforward}).

\let\oldnl\nl
\newcommand{\nlnonumber}{\renewcommand{\nl}{\let\nl\oldnl}}
\begin{algorithm}[t]

    
    \nlnonumber
    \textbf{Input: } vector loss function \( \bm{L} \), train set \( \calD \),  matrix of model parameters  \( \bm{\Theta}=\begin{bmatrix}\nnparams{1},\cdots, \nnparams{\numtasks} \end{bmatrix}^\top\),   distribution parameters \( \pp\), window \( W\in\bbN \), regularization coefficient \( \lambda \), network \( f \)
    
    Initialize each \( \nnparams{v} \) independently

     \For{batch \( (\xx, \yy) \subseteq \calD\) }{
         \( \calV\gets\varnothing \)

         \For{ \( i \in \{1,2,\dots,W\}\) }{
                 sample  \( \aa_i \sim \textrm{Dir}(\pp) \) \label{line:sampling}

             \( \calV\gets\calV\cup\aa_i \)

            \tcc{create net in ensemble's convex hull}
            \( \bmth_i\gets \aa_i ^\top  \bm{\Theta} \) \label{line:construct convex hull} 

            \( \bm{L}(\aa_i) \gets \texttt{criterion}\left(f(\xx;\bmth_i), \yy \right) \) \label{line:calculate loss}

        }
         construct multi-forward graphs \( \calG_t=(\calV,\calE_t), \forall t \) 
        \label{line:multiforward calc}

        \tcc{multiforward regularization, \Autoref{sec:algo extensions}}
        \( \calR {\gets} \sum\limits_{t=1}^{T}\log\left( \frac{1}{|\calE_t|} \sum_{(\aa_i, \aa_j)\in\calE_t} e^{\left[\calL_t(\aa_i) -\calL_t(\aa_j)\right]_+ } \right)\label{line:reg multiforward}  \) 

        \( \loss_{\textrm{total}}\gets  \sum_{i=1}^{W} \aa_i ^\top \vectorloss(\aa_i) + \lambda \cdot  \calR\)\label{line:multiforward reg}

        Backpropagate \( \loss_{\textrm{total}} \) and 
        Gradient \textit{descent} on \( \bm{\Theta} \)
    }
    \caption{\small\texttt{ParetoManifoldLearning}}
    \label{algo:full}
\end{algorithm}

\begin{claim}
    \label{claim}
    Let \( \{\bm{\theta}_t^* \in\calR: t\in [T]\}\)  be the optimal ensemble parameters retrieved at the end of training by \Autoref{algo:full} and let \( \calS \) be the their convex hull. Then \( \calS \) is a Pareto Subspace.
\looseness=-1
\end{claim}
\looseness=-1

Note that we have chosen a convex hull parameterization of the weight space, but there are other options, such as Bezier curves or other nonlinear paths \citep{Wortsman_Horton_Guestrin_etal_2021,Draxler_Veschgini_Salmhofer_etal_2018}. However, the universal approximation theorem implies no loss of generality for our design choice 
and \autoref{theorem:universal_approximation} attests to the existence of such parameterizations. 
In practice, \Autoref{claim} is validated by uniformly sampling the discovered subspace and the definition of a \textit{Pareto Subspace} is relaxed to conform to the nonconvex settings of Deep Learning, i.e., points are called Pareto optimal if the characterization holds in an open neighborhood rather than globally.
\begin{theorem}
    \label{theorem:universal_approximation}
    Given a compact $A \subset \mathbb{R}^D$ and a family of continuous
    mappings $f_n: A \rightarrow \mathbb{R}^{D'}, n = 1, \dots, N$, for
    any $\epsilon > 0$, there exists a ReLU multi-layer perceptron $f$
    with two different weight parameterizations $\bm{\theta}$ and $\bm{\theta}^\prime$, such that
    \( \forall n \in \{ 1,  \dots, N \}, \exists \alpha \in [0,1], \forall \xx \in A, \)
    \[
    \left|f_n(\xx)-f\left(\xx; \alpha \bm{\theta} + \left(1-\alpha\right) \bm{\theta}^\prime\right)\right| \leq \epsilon.
    \]
\end{theorem}
The proof is given in \autoref{appendix:proof}.
\Autoref{theorem:universal_approximation} grounds the geometrical intuition; due to overparameterization and the universal approximation theorem the Pareto Front admits a linear paramaterization.

\begin{figure*}[t]
    \centering
    \includegraphics[width=\linewidth]{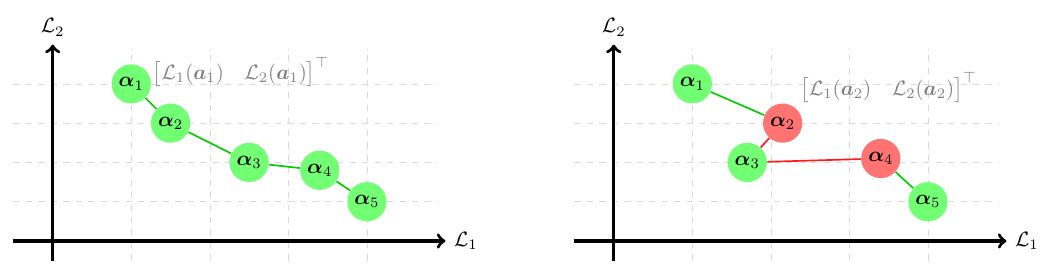}
    \caption{Visual explanation of multiforward regularization, presented in \Autoref{eq:multiforward}. The subfigures depict the loss values for various weightings \( \aa_i = [\alpha_{i,1}, \alpha_{i,2}]\). Optimal lies in the origin. We assume that \( \alpha_{1,1}>\dots>\alpha_{5,1} \). Green color corresponds to Pareto optimality. (Left) all sampled weightings are in the Pareto Front  and the regularization term is zero. (Right) The red points  are not optimal and, therefore, the regularization term penalizes the violations of the monotonicity constraints for the appropriate task loss: \( \aa_2 \) and \( \aa_4 \) violate the \( \loss_1 \) and \( \loss_2 \) orderings w.r.t. \( \aa_3 \),  since \( \alpha_{2,1}>\alpha_{3,1} \nRightarrow \calL_1(\aa_2)<\calL_1(\aa_3) \) and \( \alpha_{4,2}>\alpha_{3,2} \nRightarrow \calL_2(\aa_4)<\calL_2(\aa_3) \). }
    \label{fig:multiforwardf}
\end{figure*}

\subsection{Regularization and balancing}  
\label{sec:algo extensions}

\myparagraph{Loss and gradient balancing schemes}  
\label{sec:balancing schemes}
A common challenge in \abbrevmtl is the case where tasks have different loss scales, e.g., consider datasets with regression and classification tasks such as \utkface \citep{Zhang_ZhifeiSong_Qi_2017}. 
Then, using the same weighting \( \aa \) for both the losses and the weight ensembling, as presented in \Autoref{eq:erm-simplex}, the easiest tasks are favored and the important property of scale invariance is neglected. To prevent this, the loss weighting needs to be adjusted.
Hence, we propose simple balancing schemes: one loss and one gradient balancing scheme, whose effect is to  warp the space of loss weightings.  While gradient balancing schemes are applied on the shared parameters, loss balancing also affects the task-specific decoders, rendering the methodologies complementary. To avoid cluttering, balancing schemes are not presented in \Autoref{algo:full}.

In terms of loss balancing, we use a lightweight scheme of adding a normalization coefficient to each loss term which depends on past values. Concretely, let \( W\in\bbZ_+ \) be a positive integer and \( \calL_m( \tau_0) \) be the loss of task \( m \) in step \( \tau_0 \). Then, the regularization coefficient is \( \overline{\calL}( \tau_0; W) = \frac{1}{W}\sum_{\tau=1}^{W} \calL_m( \tau_0+1 - \tau)   \) for \(  \tau_0 \geq W \) resulting in the overall loss 
\( \calL_{total} = \aa_{ \tau_0} ^\top \widehat{\vectorloss} =    \sum_{t=1}^{\numtasks} \alpha_t \frac{\calL_t( \tau_0)}{\overline{\calL_m}( \tau_0; W)}
 \).
For gradient balancing, let \( \gg_t \) be the gradient of task \( t\in[\numtasks] \) w.r.t. the shared parameters. Previously, the update rule occurred with the overall gradient \( \gg_{total} = \aa ^\top \bm{G} = \aa ^\top \begin{bmatrix}\gg_1 & \dots & \gg_\numtasks\end{bmatrix} \). We impose a unit \( \ell_2 \)-norm for gradients and perform the update with \( \widetilde{\gg}_{total} = \aa ^\top \widetilde{\bm{G}} = \aa ^\top \begin{bmatrix}\widetilde{\gg}_1 & \dots & \widetilde{\gg}_\numtasks\end{bmatrix} \) where \( \widetilde{\gg_t} = \frac{\gg_t}{\| \gg_t \|_2} \). 
\looseness=-1

\myparagraph{Improving stability by Multi-Forward batch regularization}
\label{sec:multiforward:theory}
Consider two different weightings \( \aa_1 \) and \( \aa_2 \in \Delta_{\numtasks-1} \). Without loss of generality \( [\aa_1]_0=\alpha_1 > [\aa_2]_0=\alpha_2 \). Then, ideally, the interpolated model closer to the ensemble member for task 1 has the lowest loss on that task, i.e., we would want the ordering \( \calL_1(\aa_1) < \calL_1(\aa_2) \), and, equivalently for the other tasks. 
Furthermore, if \( \aa=\begin{bmatrix}
    1-\epsilon, \nicefrac{\epsilon}{T-1}, \dots,\nicefrac{\epsilon}{T-1}
\end{bmatrix} \), only one member essentially reaps the benefits of the gradient update and moves the ensemble towards weight configurations more suitable for one task but, perhaps deleterious for the remaining ones. Thus, we propose repeating the forward pass \( W \) times for different random weightings \( \{\aa_i\}_{i\in[W]} \), allowing the advancement of all ensemble members concurrently in a coordinated way (\Autoref{line:multiforward calc}). By performing multiple forward passes for various weightings, we achieve a lower discrepancy sequence and reduce the variance of such pernicious updates. 
\looseness=-1

We also include a regularization term, which penalizes the wrong orderings and  encourages the subspace to have Pareto properties, as in \Autoref{line:multiforward reg}. Let \( \calV \) be the set of interpolation weighs sampled in the current batch \( \calV=\{\aa_w = (\alpha_{w,1},\alpha_{w,2}, \dots, \alpha_{w,\numtasks})\in\Delta_{T-1}\}_{w\in[W]} \). 
Then each task defines the \textit{directed} graph \( \calG_t = (\calV, \calE_t) \) where \( \calE_t=\{(\aa_{i}, \aa_j)\in \calV\times \calV:\alpha_{i, t}<\alpha_{j, t} \} \). The resulting regularization is defined as: 
\begin{equation}
    \label{eq:multiforward}
    \calL_{reg} =  \sum_{t=1}^{T}\log\left( \frac{1}{|\calE_t|} \sum_{(\aa_i, \aa_j)\in\calE_t} e^{\left[\calL_t(\aa_i) -\calL_t(\aa_j)\right]_+ } \right)
\end{equation}
The current formulation of the edge set penalizes heavily the connections from vertices with low values. For this reason, we only keep one outgoing edge per node, defined by the task lexicographic order, resulting in the graph \( \calG_t^{\textrm{LEX}} = (\calV, \calE_t^{\textrm{LEX}}) \) and \( |\calE_t^{\textrm{LEX}}| =W-1,\forall t\in[T]\). Note that the regularization is convex as the sum of \textit{log-sum-exp} terms. If no violations occur, the regularization term is zero. 
\Autoref{fig:multiforwardf} offers a visual explanation of the proposed regularization.

\myparagraph{The role of sampling}
\label{sec:sampling}
\looseness=-1
Another component of \Autoref{algo:full} is the sampling imposed on the convex hull parameterization. During training, the sampling distribution dictates the loss weighting used and, hence, modulates the degree of task learning.
A natural choice is the Dirichlet distribution \( \textrm{Dir}(\pp) \) where \( \pp\in\r^\numtasks_{>0} \) are the concentration parameters, since its support is the \( \numtasks \)-dimensional simplex \( \Delta_\numtasks \).
For \( \pp=p\bm{1}_T \), the distribution is symmetric; for \( p<1 \) the sampling is more concentrated near the ensemble members, for \( p>1 \) it is near the centre and for  \( p=1 \) it corresponds to the uniform distribution. In contrast, for \( p_1\neq p_2 \) the distribution is skewed. In our experiments, we use symmetric Dirichlet distributions with \( p\geq 1 \) to guide the ensemble to representations best suited for \abbrevmtl.
\looseness=-1

\section{Experiments}

\label{sec:experiments}
We evaluate our method  on several datasets, such as \multimnist, \census, \multimnistThree, \utkface and \cityscapes, and various architectures, ranging from MultiLayer Perceptrons (MLPs) to Convolutional Neural Networks (CNNs) and Residual Networks (ResNets). Each ensemble member is initialized independently. In all experiments, the learning rate for our method is \( \nummembers \)-fold the learning rate of the baselines to counteract the fact that the backpropagation step scales the gradients by \( \nummembers^{-1} \) in expectation. 
The detailed settings used for each dataset and additional experiments are provided in the appendix.
Our overarching objective is to construct continuous weight subspaces which map to Pareto Fronts in the functional space. However, our method produces a continuum of results rather than a single point, rendering tabular presentation cumbersome. For this reason, (a) for tables we present the best-of-(sampled)-subspace results, (b) we experiment on numerous two-task datasets where plots convey the results succinctly, (c) present qualitative results on three-task datasets. 
\looseness=-1
\begin{figure*}[!t]
    \centering
    \includegraphics[width=\textwidth]{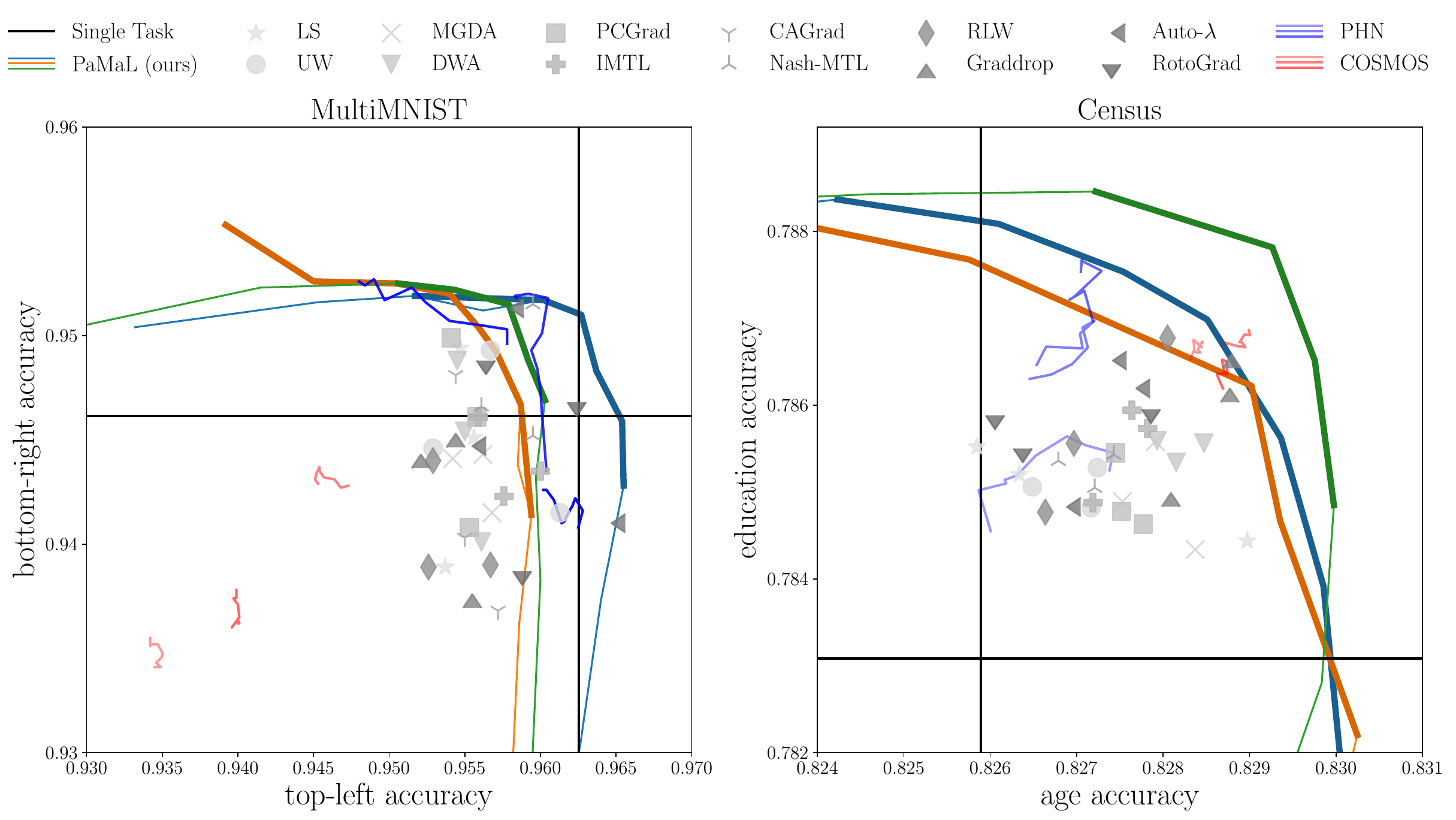}
    \caption{Experimental results on \multimnist and \census. Top right is optimal. Three random seeds per method. Solid lines correspond to our method (\pml) and thick lines to the Pareto Front. We have used a different color for each seed of \pml. Baselines are shown in shades of gray: scatter plot for MTL baselines, black lines for single task and blue/red lines for multiple-solution methods.
    In both datasets, \pmlfull discovers subspaces with diverse and Pareto-optimal solutions and outperforms the baselines.
    }
    \label{fig:small-experiments}
    \vspace*{-4px}
\end{figure*}

\def\sep{, }
\myparagraph{Baselines} We explore various algorithms from the literature:
\begin{inparaenum}[1.   ]
    \item Single-Task Learning (STL), 
    \item Linear Scalarization (LS) which minimizes the average loss \( \frac{1}{\numtasks}\sum_{t=1}^{\numtasks} \calL_t  \), 
    \item Uncertainty Weighting (UW\sep \citealt{Cipolla_Gal_Kendall_2018}),
    \item Multiple-gradient descent algorithm (MGDA\sep \citealt{Sener_Koltun_2018}), 
    \item Dynamic Weight Averaging (DWA\sep \citealt{Liu_Johns_Davison_2019}),
    \item Projecting Conflicting Gradients (PCGrad\sep \citealt{Yu_Kumar_Gupta_etal_2020}),
    \item Impartial \abbrevmtl (IMTL\sep\citealt{Liu_Li_Kuang_etal_2020}),
    \item Just Pick a Sign (Graddrop\sep \citealt{Chen_Ngiam_Huang_Luong_Kretzschmar_Chai_Anguelov_2020}),
    \item Conflict-Averse Gradient Descent (CAGrad\sep \citealt{Liu_Liu_Jin_etal_2021}),
    \item Random Loss Weighting (RLW\sep \citealt{Lin_Feiyang_Zhang_Tsang_2022}).
    \item Bargaining \abbrevmtl (Nash-MTL\sep\citealt{Navon_Shamsian_Achituve_etal_2022}),
    \item Auto-Lambda (Auto-\( \lambda \)\sep\citealt{Liu_James_Davison_Johns_2022}) and
    \item RotoGrad (\citealt{Javaloy_Valera_2022}).
\end{inparaenum} When applicable, we also explore methodologies that perform Pareto Front Approximation (PFA) in a single training run; as in
\begin{inparaenum}[1.   ]
    \setcounter{enumi}{13}
    \item Pareto HyperNetwork (PHN\sep\citealt{Navon_Shamsian_Fetaya_etal_2021}),
    \item Conditioned One-shot Multi-Objective Search (COSMOS\sep\citealt{Ruchte_Grabocka_2021}).
\end{inparaenum}
\looseness=-1

\subsection{Classification on \multimnist and \census}
\label{sec:multimnist}

We investigate the effectiveness of \pmlfull on digit classification using a LeNet model with a shared-bottom architecture and on the tabular dataset \census\cite{Kohavi_1996} for the task combination of predicting age and education level using a Multi-Layer Perceptron. The ensemble consists of two members with single task weightings.  To gauge the performance of the models lying in the linear segment between the nodes, we test the performance on the validation set on the ensemble members as well as for \( m=9 \) models uniformly distributed across the edge. We use this evaluation/plotting scheme throughout the experiments. 
We ablate the effect of multi-forward training on  \Autoref{sec:multiforward:ablation}; we use a grid search on window \( W\in\{2,3,4,5\} \) and strength \( \lambda\in\{0,2,5,10\} \) along with the base case of \( (W,\lambda)=(1,0) \) and present in the main text the setting that achieves the highest mean (across seeds) HyperVolume score on the validation set. \Autoref{fig:small-experiments} shows the results on both datasets using multi-forward regularization with window \( W=4 \) and strength \( \lambda=0 \) for \multimnist and \(W=2 \), \( \lambda=5 \) for \census.  
We observe that most baselines exhibit limited functional diversity; their predefined optimization schemes lead the differently seeded/initialized training runs to final models with similar performance (same markers are clustered in the plots). This lack of functional diversity, as well as inability to consistently outperform the Linear Scalarization baseline, are also noted by \citet{Kurin_DePalma_Kostrikov_Whiteson_Kumar_2022,Xin_Ghorbani_Garg_Firat_Gilmer_2022}.
In contrast, all \pmlfull seeds find subspaces with diverse functional solutions. This statement is quantitatively translated to higher HyperVolume compared to the baselines, shown in \Autoref{tab:hv:small experiments} of the appendix, and can be attributed to the observation that \Autoref{eq:erm-simplex} generalizes the Linear Scalarization method.

\begin{figure*}[t]
    \centering
    \def\qwidth{1} %
    \begin{subfigure}[b]{\qwidth\textwidth}
        \centering
        \includegraphics[width=\textwidth]{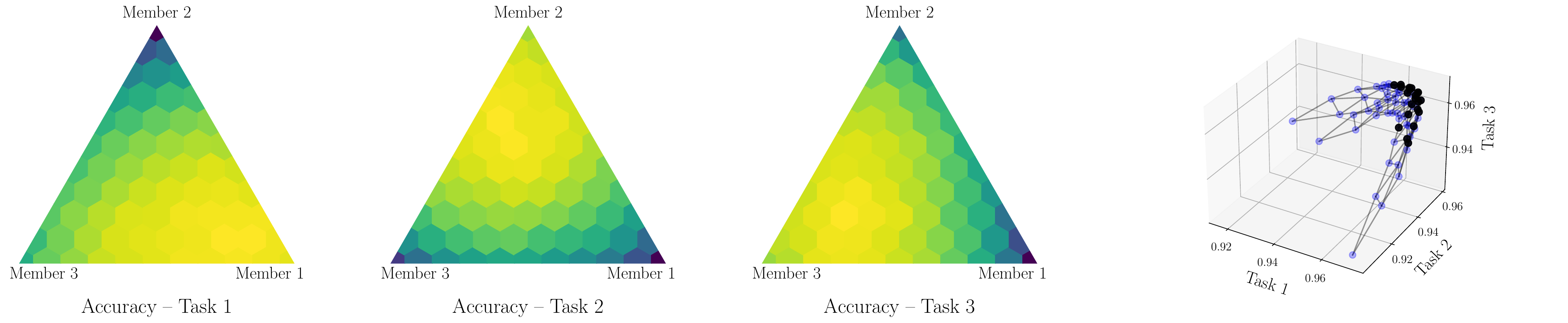}
        \caption{\multimnistThree: Accuracy Heatmap and Pareto Front for all tasks.}
        \label{fig:multimnist3maintext}
    \end{subfigure}
    \\[10pt]
    \begin{subfigure}[b]{\qwidth\textwidth}
        \centering
        \includegraphics[width=\textwidth]{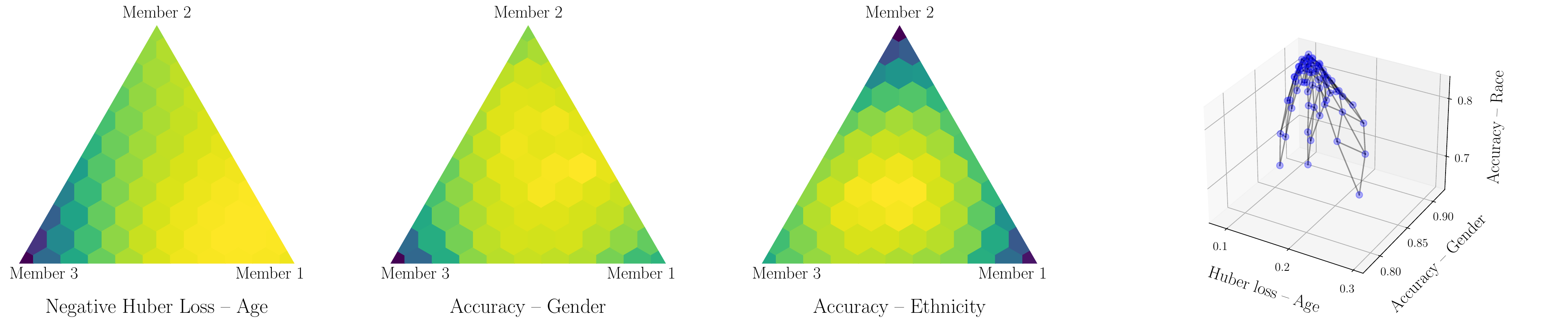}
        \caption{\utkface: Objective Heatmap and Pareto Front for all tasks.}
        \label{fig:utkface-results}
    \end{subfigure}

    \caption{Application of \pmlfull on datasets with 3 tasks. 
    Each triangle depicts the performance on a task, using color, as a function of the interpolation weighting, i.e. each hexagon corresponds to a different weighting \( \aa=[\alpha_1,\alpha_2,\alpha_3] \in \Delta_3 \). The closer the interpolated member is to a single-task predictor, the higher the performance on the corresponding task. The 3D plot, on the right, show the performance of the model in the multi-objective space. 
    }
    \label{fig:3d-task dataset}
\end{figure*}

\myparagraph{Analysis}
Task symmetry characterizes \multimnist; both digits are drawn from the same distribution, resulting in equal pace learning. However, for \census, tasks differ in statistics and, yet, the proposed method recovers a Pareto subspace with diverse solutions. 
For both datasets, we perform extensive tuning on the Pareto Front Approximation methods, i.e., PHN and COSMOS, in \autoref{sec:experimental details}. For COSMOS, the mapping from user preference to network weights is overall invalid in both datasets. Similarly, PHN produces functionally limited solutions that do not conform to the objective of Pareto optimality, while requiring \( \sim 100\times \) more parameters. These results can be attributed to two factors. First, the original papers experimented with an ample budget of \( \geq 100 \) epochs, disproportionate to the dataset complexity, and reported \textit{loss} curves instead of accuracy, neglecting the generalization gap between them. Second, the mapping from trade-off to target model via a Hypernetwork in PHN or input augmentation in COSMOS result in complex dynamics, since an \( \epsilon \)-step in the user preference is translated to different set of weights by the full forward pass of neural network.
In contrast, our approach is grounded in geometrical insights and the connection is traced back to a simple linear interpolation. In \autoref{fig:ablation for phn and cosmos} of the appendix, we supplement the ablation study for PHN and COSMOS by Spearman correlation as a proxy for monotonicity of ranking in the Pareto prism and show that these baselines must sacrifice performance in order to achieve the promise of functional diversity.

\subsection{Beyond Pairs of Classification Tasks: \multimnistThree and \utkface}  
\label{sec:beyond}

We expand the experimental validation to triplets of tasks, consider regression and more complex architectures, graduating from MLPs and CNNs to ResNets \cite{He_Zhang_Ren_etal_2016}. For three tasks, we create a 2D grid of equidistant points spanning the three single-task predictors. If \( n \) is the number of interpolated points between two (out of three) members, the grid has \( {n+1 \choose 2} \) points. We use \( n=11 \), resulting in \( 66 \) points. For visual purposes, neighboring points are connected. For three tasks, it would be visually cluttering to present the discovered subspaces with multiple seeds and baselines. Hence, we opt for a more qualitative discussion here and present quantitative findings in the appendix. 

\myparagraph{\multimnistThree} 
First, we construct an equivalent of \multimnist for 3 tasks. Digits are placed on top-left, top-right and bottom-centre. \autoref{fig:multimnist3maintext} shows the results on \multimnistThree. As argued previously, \mnist variants are characterized by task symmetry and \autoref{fig:multimnist3maintext} reflects this. For this reason, we do not employ any balancing scheme. The 3D plot in conjunction with the simplices reveal that  the method  has the effect of gradual transfer of learned representation from one member to the other, and  offers a succinct visual confirmation of \Autoref{claim}.

\myparagraph{\utkface}
The \utkface dataset \citep{Zhang_ZhifeiSong_Qi_2017} has more than 20,000 face images and three tasks: predicting age (modeled as regression using Huber loss - similar to \cite{Ma_Du_Matusik_2020}), classifying gender and ethnicity. The introduction of a regression task implies that losses have vastly different scales, which dictates the use of balancing schemes, as discussed in \Autoref{sec:algo extensions}.  We apply the proposed gradient-balancing scheme and present the results in \Autoref{fig:utkface-results}. For visual unity and to remain in the theme of ``higher is better", the \textit{negative} Huber loss is plotted. Despite the increased complexity, both in terms of network architecture and dataset, and the existence of a regression task, the proposed method discovers a \textit{Pareto Subspace}. Additional experiments and qualitative results are provided in \Autoref{sec:utkface:extra experiments}. 
\autoref{fig:3d-task dataset}, and in more detail \autoref{sec:appendix:pareto vs loss valley intersection}, show that (most of) the subspace engenders high performance and, implicitly low loss, for each task separately. Hence, the approach discovers flat regions which are linked to generalization \citep{Foret_Kleiner_Mobahi_etal_2021}. There is a also dynamic transition in the weighted (multi-task) loss landscape w.r.t. weight \( \aa \), which leads to the desired Pareto properies.

\vspace*{-4px}
\subsection{Scene understanding}  
\label{sec:scene understanding}

\looseness=-1

We also explore the applicability of \pmlfull for \cityscapes\citep{Cordts_Omran_Ramos_etal_2016}, a scene understanding dataset containing high-resolution images of urban street scenes. Our experimental configuration is drawn from \cite{Liu_Johns_Davison_2019,Yu_Kumar_Gupta_etal_2020,Liu_Liu_Jin_etal_2021,Navon_Shamsian_Achituve_etal_2022} with some modifications. Concretely, we address two tasks: semantic segmentation and depth regression. We use a SegNet architecture \citep{Badrinarayanan_Kendall_Cipolla_2017} trained for 100 epochs with Adam optimizer \citep{Kingma_Ba_2015} of initial learning rate \( 10^{-4} \), which is halved after 75 epochs. The images are resized to \( 128\times256 \) pixels. 
We use 500 of the 2975 training images for validation, and report the test results in \autoref{tab:cityscapes}. 
We use gradient balancing, window \( W=3 \) and \( \lambda=1 \), while the concentration parameter of the Dirichlet distribution is set to \( p=7 \), helping convergence. Additional results are presented in \autoref{sec:appendix:cityscapes:extra experiments}.
In Depth Estimation and out of MTL methods, \pmlfull is optimal along with MGDA (lower is better). In Semantic Segmentation (higher is better), however, MGDA performs poorly and is clearly dominated by all methods, while our approach is outperforming most baselines and offers a balanced solution. Compared to the multi-solution baselines, COSMOS showcases task bias performing poorly on Depth Estimation, while PHN is ommitted altogether due to not scaling to large networks.
It is remarkable that, despite our goal of discovering \textit{Pareto subspaces}, the proposed method is dominating most of the state-of-the-art algorithms, attesting to the flexibility of the weight ensembles in \mtl. 
\newcommand{\myfontsize}[1]{\begin{scriptsize}#1\end{scriptsize}}
\setlength{\tabcolsep}{5pt}

\begin{table}[t]
\footnotesize
\renewcommand{\arraystretch}{0.95}
\centering
\caption{Test performance on \textit{CityScapes}. 3 random seeds per method. For \pmlfull, we report the mean (across seeds) best results from the final subspace. Methods are divided into single-task, single-solution \abbrevmtl, multi-solution \abbrevmtl and proposed method.} 
\begin{tabular}{@{}lccccS[table-format=3.2]cccc@{}}
\toprule
  & \multicolumn{2}{c}{Segmentation} &                      & \multicolumn{2}{c}{Depth} \\
	\cmidrule(lr){2-3} \cmidrule(lr){5-6}
   & \multicolumn{1}{l}{mIoU $\uparrow$} & \multicolumn{1}{l}{Pix Acc $\uparrow$} & \multicolumn{1}{l}{} & \multicolumn{1}{l}{Abs Err $\downarrow$} & \multicolumn{1}{l}{Rel Err $\downarrow$} \\
\midrule
STL                 &      70.96 &        92.12 &                      &     \bf 0.0141 &     \bf 38.644 \\
\midrule
LS                  &      70.12 &        91.90 &                      &         0.0192 &        124.061 \\
UW                  &      70.20 &        91.93 &                      &         0.0189 &        125.943 \\
MGDA                &      66.45 &        90.79 &                      &     \bf 0.0141 &         53.138 \\
DWA                 &      70.10 &        91.89 &                      &         0.0192 &        127.659 \\
PCGrad              &      70.02 &        91.84 &                      &         0.0188 &        126.255 \\
IMTL                &      70.77 &        92.12 &                      &         0.0151 &         74.230 \\
Graddrop            &      70.07 &        91.93 &                      &         0.0189 &        127.146 \\
CAGrad              &      69.23 &        91.61 &                      &         0.0168 &        110.139 \\
RLW                 &      68.79 &        91.52 &                      &         0.0213 &        126.942 \\
Nash-MTL            &  \bf 71.13 &    \bf 92.23 &                      &         0.0157 &         78.499 \\
RotoGrad            &      69.92 &        91.85 &                      &         0.0193 &        127.281 \\
Auto-$\lambda$      &      70.47 &        92.01 &                      &         0.0177 &        116.959 \\
\midrule
COSMOS              &      69.78 &        91.79 &                      &         0.0539 &        136.614 \\
\midrule
PaMaL(ours) &     70.35 &        91.99 &                      &     \bf 0.0141 &         54.520 \\
\bottomrule
\end{tabular}

\label{tab:cityscapes}
\vspace*{-10px}
\end{table}
\looseness=-1

\section{Conclusion}  
\label{sec:conclusion}
In this paper, we proposed a weight-ensembling method tailored to \mtl; multiple single-task predictors are trained in conjunction to produce a subspace formed by their convex hull, and endowed with desirable Pareto properties. We experimentally show on a diverse suite of benchmarks that the proposed method is successful in discovering \textit{Pareto subspaces} and outperforms or is on par with state-of-the-art MTL methods. An interesting future direction is to perform a hierarchical weight ensembling, sharing progressively more of the lower layers, given that the features learned at low depth are similar across tasks.
An alternative exploration venue is to connect our method to the challenge of task affinity \citep{fifty2021efficiently,Standley_Zamir_Chen_etal_2020} via a geometrical lens of the loss landscape.

\section*{Acknowledgments}  
The work of Nikolaos Dimitriadis was supported by Swisscom (Switzerland) AG. We would like to thank Guillermo Ortiz-Jiménez, Apostolos Modas, Cl\'{e}ment Vignac, Prabhu Teja, and the anonymous reviewers for their valuable feedback.

\newpage
\onecolumn
\twocolumn
\bibliography{a}
\bibliographystyle{icml2023}

\clearpage
\appendix

\showappendixtrue
\onecolumn
\section*{Appendix Overview}
\label{sec:appendix overview}

As a reference, we provide the following table of contents solely for the appendix.

\begin{enumerate}[label=\Alph*.]
    \item Discussion 
    \begin{enumerate}[label=A\arabic*.]
        \item Effect of sampling on the Pareto properties of the discovered subspace\dotfill\Autoref{sec:appendix:sampling}
        \item Connection between Pareto Optimality and multiple valley intersections\dotfill\Autoref{sec:appendix:pareto vs loss valley intersection}
        \item Proof of \autoref{theorem:universal_approximation}\dotfill\Autoref{appendix:proof}
    \end{enumerate}
    \item Ablations
    \begin{enumerate}[label=B\arabic*.]
        \item Ablation on Multi-Forward Regularization\dotfill\Autoref{sec:multiforward:ablation}
        \item Illustrative example: ablation on loss/gradient balancing schemes\dotfill\Autoref{sec:illustrative details}
        \item \utkface: ablation on the effect of loss/gradient balancing schemes\dotfill\Autoref{sec:utkface:extra experiments}
        \item Hyperparameter optimization for PHN and COSMOS\dotfill\Autoref{sec:appendix:phn and cosmos ablation}
    \end{enumerate} 
    \item Additional Experiments
    \begin{enumerate}[label=C\arabic*.]
        \item Details on experimental configurations\dotfill\Autoref{sec:experimental details}
        \item HyperVolume analysis on \multimnist and \census\dotfill\Autoref{sec:hv analysis}
        \item \multimnistThree quantitative results\dotfill\Autoref{sec:multimnist3:extra experiments} 
        \item \cityscapes additional results\dotfill\Autoref{sec:appendix:cityscapes:extra experiments}
    \end{enumerate}
\end{enumerate}

\ifshowappendix
        \conditionalclearpage
        \section{Discussion}
        \label{appendix:discussion}
        \subsection{Effect of sampling on the Pareto properties of the discovered subspace}
\label{sec:appendix:sampling}  
{

\begin{figure*}[!t]
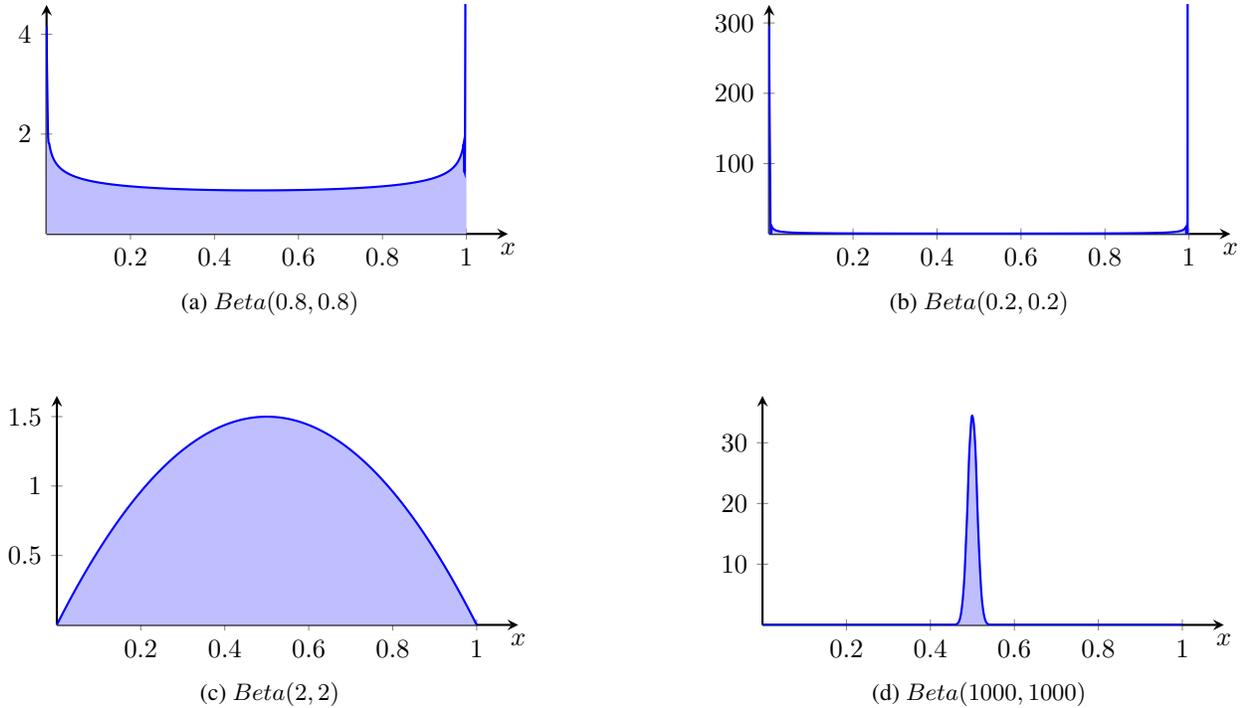

\centering
\begin{subfigure}[b]{0.45\textwidth}
    \centering
    \plotbeta{0.8}{0.8}{blue}
    \caption{\( Beta(0.8, 0.8) \)}
\end{subfigure}
\hfill
\begin{subfigure}[b]{0.45\textwidth}
    \centering
    \plotbeta{0.2}{0.2}{blue}
    \caption{\( Beta(0.2,0.2) \)}
\end{subfigure}
\\[1cm]
\begin{subfigure}[b]{0.45\textwidth}
    \centering
    \plotbeta{2}{2}{blue}
    \caption{\( Beta(2,2) \)}
\end{subfigure}
\hfill
\begin{subfigure}[b]{0.45\textwidth}
    \centering
    \plotbeta{1000}{1000}{blue}
    \caption{\( Beta(1000,1000) \)}
\end{subfigure}
\caption{Dirichlet distribution in the case of two tasks. Top row: \( p<1 \) and the distribution is more concentrated towards the ensemble members. Bottom row: \( p>1 \) and the distribution focuses more on the midpoint which corresponds to all tasks having the same weight. Right column: extreme choices \( p \rightarrow 0 \) or \( p \rightarrow \infty  \). Left column: milder choices.}
\label{fig:distributions}
\end{figure*}

This appendix expands on \Autoref{sec:algo extensions} and, specifically, presents in greater detail the intuition behind the sampling distribution's parameters. 
Let  \( \pp\in\r^\numtasks_+ \) be the parameters of the Dirichlet distribution. Assuming no prior knowledge on the tasks, e.g., task difficulties or affinities, a symmetric distribution is used by setting \( \pp=p\mathbf{1}_T \). This design choice results in three cases:
\begin{itemize}
    \item \( p=1 \): the distribution is uniform on the simplex. Intuitively this means that all tasks are equally important and we care about the diversity of solutions for all tradeoffs (reflected in the linear scalarization weights).
    \item \( p\in(0,1) \): the distribution is more concentrated towards the ensemble members, as in the top row of \Autoref{fig:distributions}. Assume an extreme case of two tasks and \( p=0 \). Then the distribution degenerates to a Bernoulli distribution. Effectively, at each iteration one of the ensemble members is selected and its weights are updated, which will result in two separate and independent single-task predictors with no common representation infused about the other task. Then, linearly interpolating in weight space will result in models with random predictions for both tasks, since the training procedure has not focused in retrieving a Pareto Subspace. 
    
    For milder cases (e.g. \( p=0.7 \)) , we observed that the models in the middle of the linear interpolation suffered in performance which can be attributed to the fact that the sampling focused more on single-task rather than multi-task representations and performance.
    \item \( p>1 \). Then the distribution is more concentrated towards the midpoint of the simplex, as in the bottom row of \Autoref{fig:distributions}. Assume an extreme case of two tasks and \( p \rightarrow \infty  \). Then, the distribution becomes deterministic and outputs equal weights for all tasks. The randomly and independently initialized ensemble members will collapse to each other, resulting in duplicate ensemble members. Similarly, for very large values (e.g. \( p=100 \)), the functional diversity of the ensemble will suffer since the weights produced by the distribution will be almost equal for all tasks, resulting in a milder version of the aforementioned phenomenon. In contrast, we found that small values such as \( p=2 \) or \( p=3 \) can help convergence since they put more emphasis towards common representation (compared to \( p=1 \)), but may limit functional diversity.
\end{itemize}

\Autoref{fig:distributions:experiments} presents experimental results on \multimnist and \census for various concentration parameters \( p\in\{0, 0.1, 100\} \) of the Dirichlet distribution. Let \( \nnparams{1} \) and \( \nnparams{2} \) be the parameters of the ensemble members. For \( p=0 \), the ensemble consists of two single-task predictors with no multitask learning representational knowledge, since their interpolation meets a low accuracy/high loss barrier. We omit the case of \( p=0 \) for \census for visual clarity. This lack of common representation is evident in the cosine similarities as well, where for \( p=0 \) \( \cos(\nnparams{1}, \nnparams{2})\approx 0 \). On the other hand, for \( p=0.1 \), common representations are infused into the ensemble and the experimental results show that the test performance is characterized by diversity. However, this comes at the expense of the interpolated models at the middle of the line segment, where the performance is suboptimal compared to \( p=100 \) for \multimnist. This behavior is also illustrated in the cosine similarities, where for \( p=100 \) the ensemble weights \( \aa \) are in an \( \epsilon \)-ball around the midpoint causing the independently initialized models to progressively collapse. For \( \census \), we also observe that this collapsing leads to very high cosine similarity \( \cos(\nnparams{1}, \nnparams{2}) > 0.9 \) and the ensemble is suboptimal compared to \( p=0.1 \).

\begin{figure*}[!t]
    \centering
    \begin{subfigure}[b]{0.45\textwidth}
        \centering
        \includegraphics[width=\linewidth]{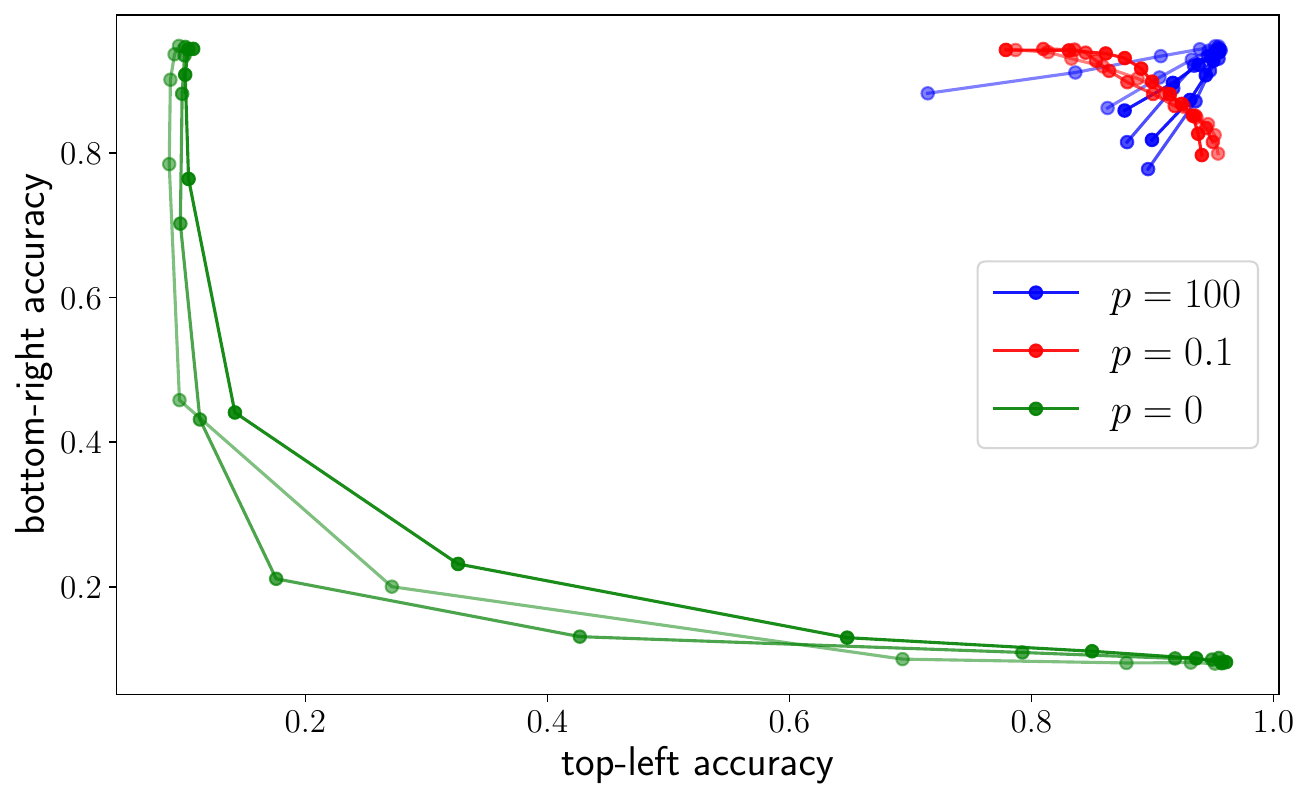}
        \caption{\multimnist: Experimental results using three random seeds per method.}
    \end{subfigure}
    \hfill
    \begin{subfigure}[b]{0.45\textwidth}
        \centering
        \includegraphics[width=\linewidth]{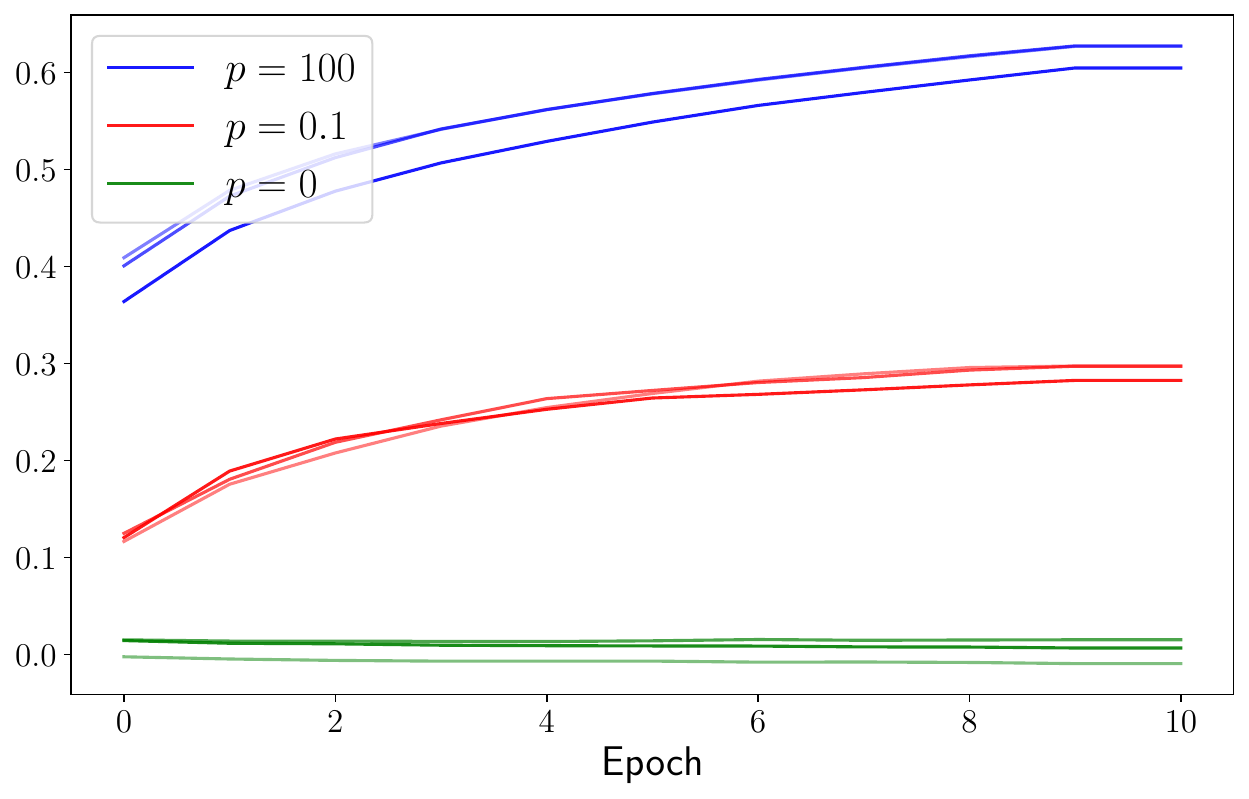}
        \caption{\multimnist: Cosine similarities of ensemble members.}
    \end{subfigure}
    \\[1cm]
    \begin{subfigure}[b]{0.45\textwidth}
        \centering
        \includegraphics[width=\linewidth]{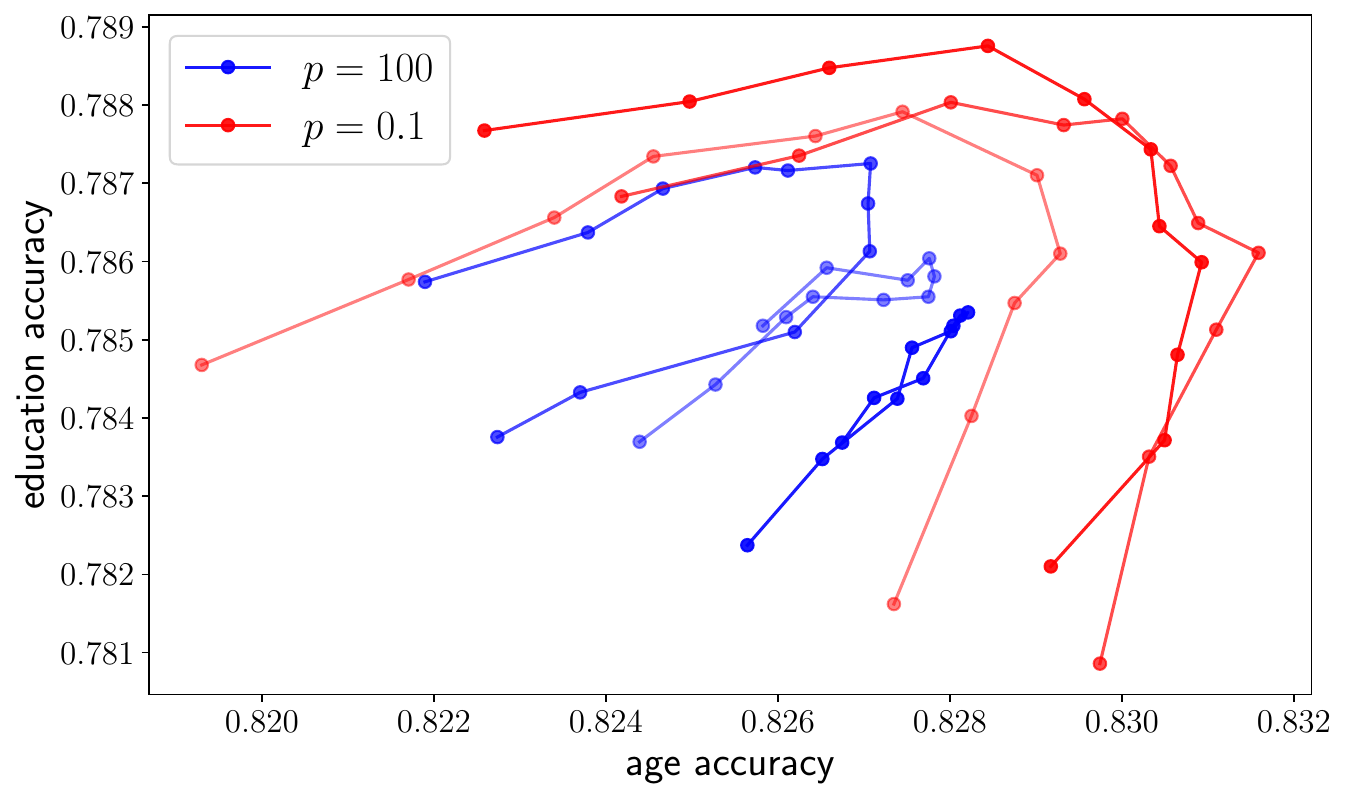}
        \caption{\census: Experimental results using three random seeds per method.}
    \end{subfigure}
    \hfill
    \begin{subfigure}[b]{0.45\textwidth}
        \centering
        \includegraphics[width=\linewidth]{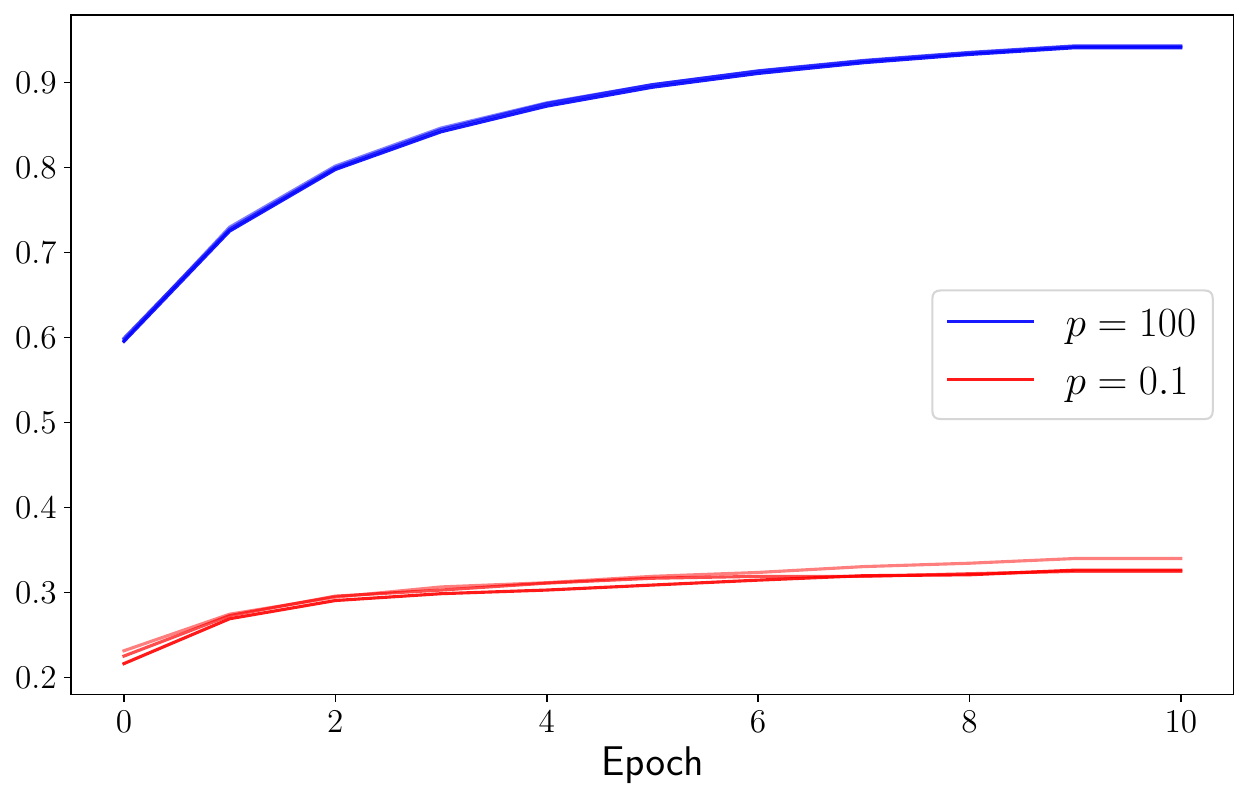}
        \caption{\census: Cosine similarities of ensemble members.}
    \end{subfigure}

    \caption{Experimental results on \multimnist and \census varying the concentration parameters \( \pp=p\bm{1}_\numtasks \) of the sampling distribution. Three seeds depicted in shades of the same colors for the various \( p \).}
    \label{fig:distributions:experiments}
\end{figure*}

}
        \subsection{Connection between Pareto Optimality and multiple valley intersections}
\label{sec:appendix:pareto vs loss valley intersection}  
{

In this section, we investigate the connection between the intersection of multiple loss landscapes, pareto optimality and the effect of the proposed algorithm Pareto Manifold Learning. We use the illustrative example, presented in \Autoref{fig:illustrative}. Let \( \bm{\Theta} \) be the parameter space of the model and \( \loss_t : \bm{\Theta}\rightarrow \r, t\in\{1,2\}    \), be the losses of the problem. For \( \alpha\in[0,1] \) and \( \bm{\theta}\in\bm{\Theta} \), the overall objective is \( \loss(\bm{\theta}, \alpha)=\alpha \loss_1(\bm{\theta}) +(1- \alpha) \loss_2(\bm{\theta}) \). 

\Autoref{fig:pareto appendix:color} presents the overall loss objective as \( \alpha \) varies from 0 to 1. For the extreme values of the range, the loss landscape is inherently single-task. The subspace discovered by the method is depicted in blue, while a black `x' is used for the corresponding interpolated model, i.e., it corresponds to \( \loss(\alpha\nnparams{1}+(1-\alpha)\nnparams{2}, \alpha) \).  
In other words, the proposed method tracks the optimum in parameter space as the overall objective evolves and the various loss landscapes are weighted accordingly. While an acceptable multi-task solution lies in the  intersection of low loss landscapes, Pareto Manifold Learning focuses on the aforementioned dynamic scenario of loss weighting.  

\begin{figure*}[!h]
    \centering
    \includegraphics[width=\textwidth]{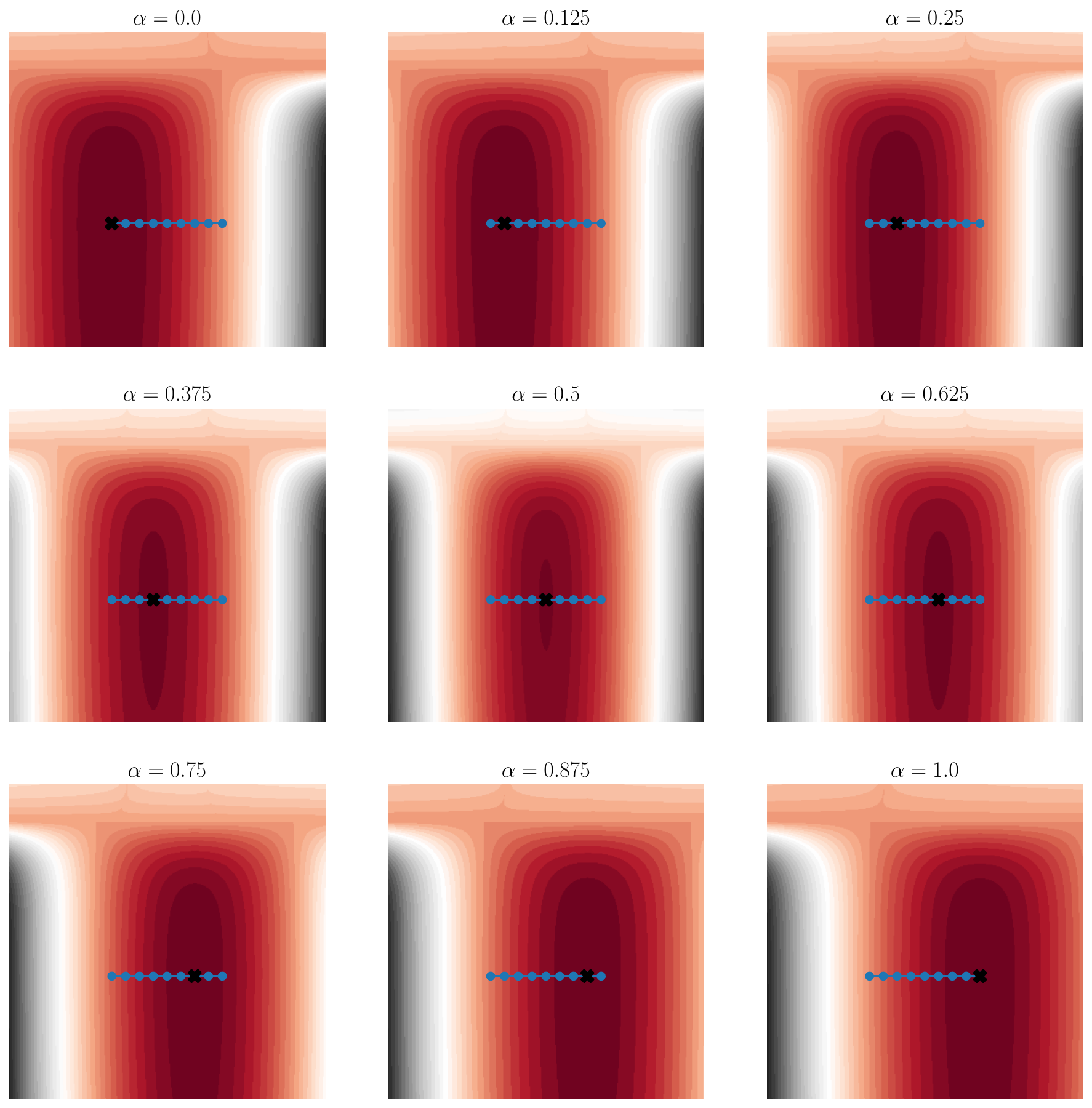}
    \caption{\textit{Illustrative example}: (Overall) loss surface as a function of the model's weights. The overall objective is \( \loss(\bm{\theta}, \alpha)=\alpha \loss_1(\bm{\theta}) +(1- \alpha) \loss_2(\bm{\theta})\) and is shown for various values of \( \alpha \). The Pareto subspace discovered by the proposed method is depicted in blue. `X' shows the solution of the method for the corresponding \( \alpha \).}
    \label{fig:pareto appendix:color}
\end{figure*}

}
        \clearpage
        
\subsection{Proof of \autoref{theorem:universal_approximation}}
\label{appendix:proof}
\def\relu{\sigma}

The theorem below shows that given a family of mappings, we can approximate them [arbitrarily accurately] with linear interpolation of two perceptrons in parameter space.

\paragraph*{Theorem} \textit{Given a compact $A \subset \mathbb{R}^D$ and a family of continuous
mappings $f_n: A \rightarrow \mathbb{R}^{D'}, n = 1, \dots, N$, for
any $\epsilon > 0$, there exists a ReLU multi-layer perceptron $f$
with two different weight parameterizations $\bm{\theta}$ and $\bm{\theta}^\prime$, such that
\[
  \forall n \in \{ 1,  \dots, N \}, \exists \alpha \in [0,1], 
\forall \xx \in A, \left|f_n(x)-f\left(\xx; \alpha \bm{\theta} + \left(1-\alpha\right) \bm{\theta}^\prime\right)\right| \leq \epsilon
\]
}
\vspace*{-15px}

\begin{proof}

  Let $\relu$ be the ReLU non-linearity $x \mapsto \max(0,x)$.
  
  From the universal representation theorem, there exists $Q \in
  \mathbb{N}, \bm{M} \in \mathbb{R}^{(D+1) \times Q}, \bm{B} \in \mathbb{R}^Q,
  \bm{M'} \in \mathbb{R}^{Q \times D'}$ such that with the one
         hidden layer perceptron
  \begin{align*}
  g : A \times [0,1] & \rightarrow \mathbb{R}^{D'} \\
      \bm{z}              & \mapsto \bm{M'} \relu(\bm{M} \bm{z} + \bm{B}),
  \end{align*}
  we have
  \[
  \forall \bm{x} \in A, \forall n \in \{ 1,  \dots, N \}, \left|f_n(\bm{x})-g\left(x_1, \dots, x_D, \frac{n-1}{N-1}\right)\right| \leq \epsilon.
  \]
  
  Let
  \[
    \bm{R} = \left(
  \begin{array}{rrrrrr}
   1 & 0      & 0      & 0  &        & 0                            \\
  -1 & 0      & 0      & 0  &        & 0                            \\
   0 & 1      & 0      & 0  &        & 0                            \\
   0 & -1     & 0      & 0  & \ldots & 0                            \\
   0 & 0      & 1      & 0  &        & 0                            \\
   0 & 0      & -1     & 0  &        & 0                            \\
     & \vdots &        &    &        & \vdots                       \\
   0 & 0      & 0      & 0  &        & 1                            \\
   0 & 0      & 0      & 0  & \ldots & -1                           \\
   0 & 0      & 0      & 0  &        & 0
  \end{array}
  \right)
  \quad \text{and} \quad
  \bm{S} = \left(
  \begin{array}{rrrrrrrrrrr}
   1 & -1     & 0      & 0  & 0      & 0  &        & 0 & 0      & 0 \\
   0 & 0      & 1      & -1 & 0      & 0  & \ldots & 0 & 0      & 0 \\
  0  & 0      & 0      & 0  & 1      & -1 &        & 0 & 0      & 0 \\
     &        & \vdots &    &        &    &        &   & \vdots &   \\
   0 & 0      & 0      & 0  & 0      & 0  & \ldots & 1 & -1     & 0 \\
   0 & 0      & 0      & 0  & 0      & 0  &        & 0 & 0      & 1
  \end{array}
  \right),
  \]
  and let $\bm{U_k} = (\underbrace{0, \dots, 0}_{\times 2D},k)$.
  
  Then, with $\bm{x} \in \mathbb{R}^D$, we have
  \[
  \forall \alpha \geq 0, \quad \bm{S} \relu(\bm{R} x + \alpha \bm{U_1} + (1-\alpha) \bm{U_0}) = (x_1, \dots, x_D, \alpha).
  \]
  
  So with $\bm{\theta}=(\bm{R}, \bm{U_1}, \bm{M S}, \bm{B}, \bm{M'})$ and $\bm{\theta}'=(\bm{R}, \bm{U_0}, \bm{M S}, \bm{B},\bm{M'})$ and
  \[
  f(\bm{x;r,u,m,b,m'}) = \bm{m'} \relu ( \bm{m} \relu (\bm{rx + u}) + \bm{b}),
  \]
  then
  \begin{align*}
  f(x;\alpha \bm{\theta} + (1-\alpha) \bm{\theta}') & = f(\xx; \bm{R}, \alpha \bm{U_1} + (1-\alpha)\bm{U_0}, \bm{M S}, \bm{B}, \bm{M}')           \\
                                   & = \bm{M'} \relu ( \bm{M S }\relu (\bm{R x} + \alpha\bm{ U_1} + (1-\alpha)\bm{U_0}) + \bm{B}) \\
                                   & = g(\bm{S} \relu (\bm{R x} + \alpha \bm{U_1} + (1-\alpha)\bm{U_0}))                \\
                                   & = g(x_1, \dots, x_D, \alpha)
  \end{align*}
  
  \end{proof}
  
        \conditionalclearpage

        \section{Ablation Studies}
        \label{appendix:ablations}
        \subsection{Ablation on Multi-Forward Regularization}  
\label{sec:multiforward:ablation}

\begin{figure*}[t]
    \centering
    \includegraphics[width=0.95\textwidth]{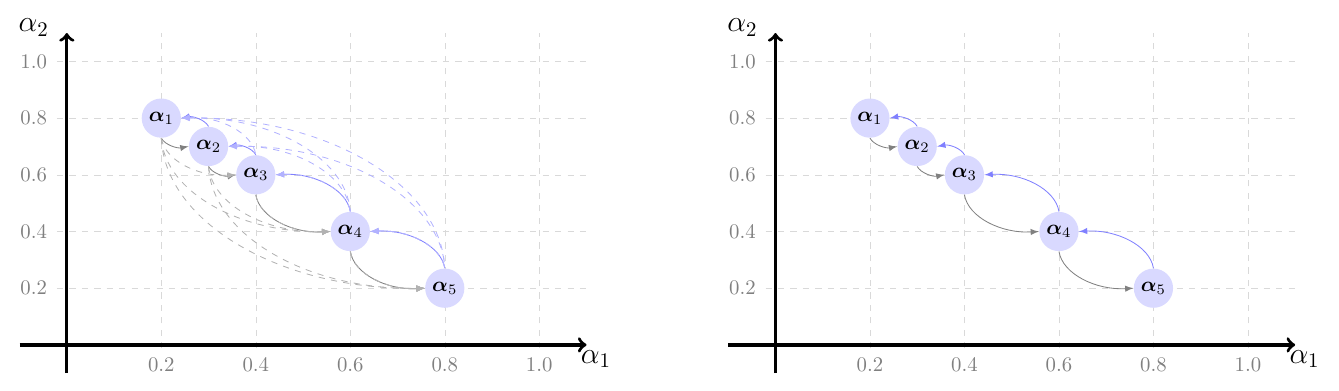}
    \caption{Multi-Forward Graph: case of two tasks. We assume a window of \( W=5 \). The nodes lie in the line segment \( \alpha_2+\alpha_1=1 \), \( \alpha_1,\alpha_2\in[0,1] \). (Left) Full graph and dashed edges will be removed. (Right) Final graph. }
    \label{fig:multiforward:case of 2 tasks}
\end{figure*}

\begin{figure*}[t]
    \centering
    \includegraphics[width=0.95\textwidth]{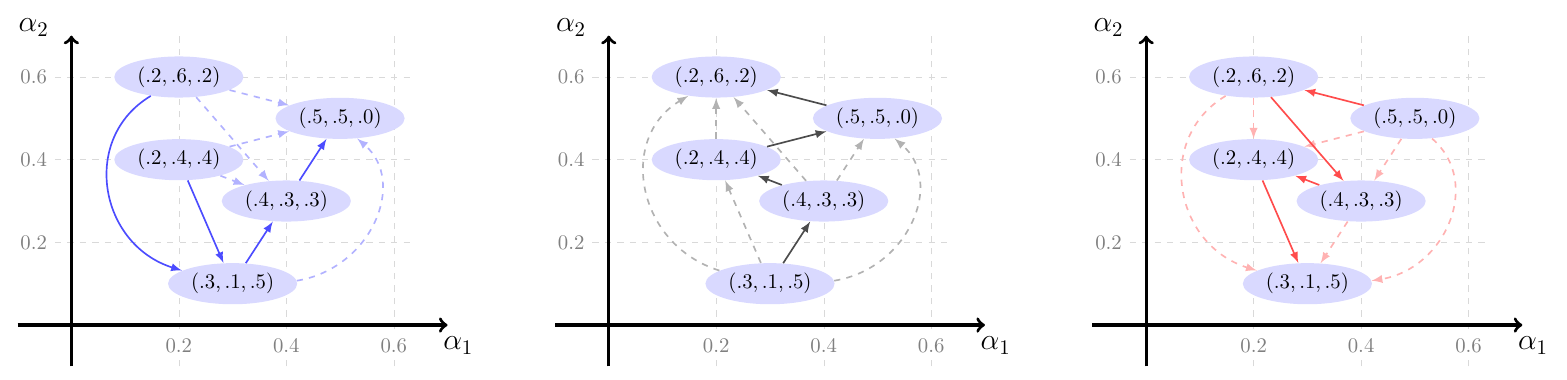}
    \caption{Multi-Forward Graph for three tasks. Left, middle and right present the case of the first, second and third task, respectively. Each node is noted by its weighting, summing up to 1. Edges are drawn if the two nodes obey the total ordering imposed by the task. Dashed edges are omitted from the final graph.
    }
    \label{fig:multiforward:case of 3 tasks}
\end{figure*}

Multi-Forward regularization, introduced in \Autoref{sec:algo extensions}, penalizes the ensemble if the interpolated models' losses (sampled within a batch) are not in accordance with the tradeoff imposed by the corresponding interpolation weights. Simply put, the closer we sample to the member corresponding to task 1, the lower the loss should be on task 1. The same applies to the other tasks. \Autoref{fig:multiforwardf} presents the case of two tasks, where the idea of the regularization is outlined in loss space. For completeness, we present the underlying graph construction for the cases of two and three tasks in \Autoref{fig:multiforward:case of 2 tasks} and \Autoref{fig:multiforward:case of 3 tasks}, respectively. The nodes of the graphs are associated with the sampled weightings and the edges for the graph \( \calG_t \) of task \( t \) are drawn w.r.t. the corresponding partial ordering. If the loss ordering is violated for a given edge, a penalty term is added. 

We ablate the effect multi-forward training and the corresponding regularization have on performance. We explore the \multimnist and \census datasets using the same experimental configurations as in the main text. We are interested in:
\begin{itemize}[topsep=0pt,itemsep=-1pt]
    \item \( W \): number of \( \aa \) re-samplings per batch. This parameter is also referred as \textit{window}.
    \item \( \lambda\): the regularization strength as presented in \Autoref{algo:full}. For \( \lambda=0 \), no regularization is applied but the subspace is still sampled \( W \) times and the total loss takes into account all the respective interpolated models.
\end{itemize}
\Autoref{fig:multimnist-multiforward} and \Autoref{tab:hv analysis:multimnist} present the results for \multimnist.  \Autoref{fig:census-multiforward} and \Autoref{tab:hv analysis:census} present the results for \census. 
It is important to note that \multimnist is symmetric, while \census is not. 
As a result, the features learned for each single-task predictor are helpful to one another and the case of \( \lambda=0 \), i.e., no regularization and only multi-forward training, is beneficial for \multimnist but not for \census. Intuitively, both digit classification tasks have the same difficulty and posterior distribution, which produces few violations of monotonicity constraints and renders the regularization less applicable. On the other hand, severe regularization such as \( \lambda=10 \) can be harmful and hinder training. More details in table and figure captions.

\begin{table*}[t]
    \centering
    \caption{\multimnist: Ablation on multi-forward training and regularization, presented in \Autoref{sec:algo extensions}. Validation performance in terms of HyperVolume (HV) metric. Higher is better, except for standard deviation (std). The visual complement of the table appears in \Autoref{fig:multimnist-multiforward}. For each configuration, we track the Hypervolume across three random seeds and present Mean HV, max HV and standard deviation. We annotate with bold the best per column. In the main text, we report the best result in terms of mean HV, i.e., \( W=4 \) and \( \lambda=0 \).}
    \label{tab:hv analysis:multimnist}
    \begin{tabular}{llrrrrrr}
        \toprule
              &              &          Seed - 0 &          Seed - 1 &          Seed - 2 &           Mean HV &            Max HV &               std \\
        \midrule
        $W=2$ & $\lambda=0$ &            0.9205 &            0.9083 &            0.9100 &            0.9129 &            0.9205 &            0.0054 \\
              & $\lambda=2$ &            0.9121 &            0.9105 &            0.9037 &            0.9088 &            0.9121 &            0.0036 \\
              & $\lambda=5$ &            0.9132 &            0.9016 &            0.8979 &            0.9043 &            0.9132 &            0.0065 \\
              & $\lambda=10$ &            0.8766 &            0.8932 &            0.8470 &            0.8723 &            0.8932 & 0.0191 \\
        \midrule
        $W=3$ & $\lambda=0$ &            0.9215 &            0.9141 &            0.9111 &            0.9156 &            0.9215 &            0.0044 \\
              & $\lambda=2$ &            0.9176 &            0.9150 &            0.9122 &            0.9149 &            0.9176 &            0.0022 \\
              & $\lambda=5$ &            0.9155 &            0.9138 &            0.9140 &            0.9144 &            0.9155 &            $\mathbf{0.0008}$ \\
              & $\lambda=10$ &            0.9122 &            0.9050 &            0.8962 &            0.9045 &            0.9122 &            0.0066 \\
        \midrule
        $W=4$ & $\lambda=0$ & $\mathbf{0.9220}$ &            0.9187 &            0.9143 & $\mathbf{0.9184}$ & $\mathbf{0.9220}$ &            0.0032 \\
              & $\lambda=2$ &            0.9213 &            0.9149 & $\mathbf{0.9157}$ &            0.9173 &            0.9213 &            0.0028 \\
              & $\lambda=5$ &            0.9158 &            0.9139 &            0.9132 &            0.9143 &            0.9158 &            0.0011 \\
              & $\lambda=10$ &            0.9177 &            0.9022 &            0.9102 &            0.9100 &            0.9177 &            0.0063 \\
        \midrule
        $W=5$ & $\lambda=0$ &            0.9131 &            0.9180 &            0.9156 &            0.9156 &            0.9180 &            0.0020 \\
              & $\lambda=2$ &            0.9158 & $\mathbf{0.9203}$ &            0.9146 &            0.9169 &            0.9203 &            0.0024 \\
              & $\lambda=5$ &            0.9138 &            0.9082 &            0.9140 &            0.9120 &            0.9140 &            0.0027 \\
              & $\lambda=10$ &            0.9165 &            0.9158 &            0.9121 &            0.9148 &            0.9165 &            0.0019 \\
        \bottomrule
        \end{tabular}
        
\end{table*}

\begin{table*}[t]
    \centering
    \caption{\census: Ablation on multi-forward training and regularization, presented in \Autoref{sec:algo extensions}. Validation performance in terms of HyperVolume (HV) metric. Higher is better, except for standard deviation (std). The visual complement of the table appears in \Autoref{fig:census-multiforward}. For each configuration, we track the Hypervolume across three random seeds and present Mean HV, max HV and standard deviation. We annotate with bold the best per column. In the main text, we report the best result in terms of mean HV, i.e., \( W=2 \) and \( \lambda=5 \). }
    \label{tab:hv analysis:census}
    \begin{tabular}{llrrrrrr}
        \toprule
              &              &          Seed - 0 &          Seed - 1 &          Seed - 2 &           Mean HV &            Max HV &               std \\
        \midrule
        $W=2$ & $\lambda=0$ &            0.6517 &            0.6530 &            0.6532 &            0.6526 &            0.6532 &            0.0006 \\
              & $\lambda=2$ &            0.6575 &            0.6564 &            0.6560 &            0.6566 &            0.6575 &            0.0006 \\
              & $\lambda=5$ & $\mathbf{0.6577}$ & $\mathbf{0.6574}$ & $\mathbf{0.6590}$ & $\mathbf{0.6581}$ & $\mathbf{0.6590}$ &            0.0007 \\
              & $\lambda=10$ &            0.6548 &            0.6557 &            0.6554 &            0.6553 &            0.6557 &            $\mathbf{0.0004}$ \\
        \midrule
        $W=3$ & $\lambda=0$ &            0.6517 &            0.6496 &            0.6501 &            0.6505 &            0.6517 &            0.0009 \\
              & $\lambda=2$ &            0.6540 &            0.6523 &            0.6544 &            0.6536 &            0.6544 &            0.0009 \\
              & $\lambda=5$ &            0.6552 &            0.6539 &            0.6536 &            0.6542 &            0.6552 &            0.0007 \\
              & $\lambda=10$ &            0.6574 &            0.6567 &            0.6566 &            0.6569 &            0.6574 &            $\mathbf{0.0004}$ \\
        \midrule
        $W=4$ & $\lambda=0$ &            0.6488 &            0.6516 &            0.6504 &            0.6503 &            0.6516 &            0.0011 \\
              & $\lambda=2$ &            0.6492 &            0.6522 &            0.6504 &            0.6506 &            0.6522 &            0.0012 \\
              & $\lambda=5$ &            0.6499 &            0.6514 &            0.6525 &            0.6513 &            0.6525 &            0.0011 \\
              & $\lambda=10$ &            0.6529 &            0.6549 &            0.6558 &            0.6545 &            0.6558 &            0.0012 \\
        \midrule
        $W=5$ & $\lambda=0$ &            0.6497 &            0.6502 &            0.6484 &            0.6494 &            0.6502 &            0.0008 \\
              & $\lambda=2$ &            0.6478 &            0.6497 &            0.6495 &            0.6490 &            0.6497 &            0.0009 \\
              & $\lambda=5$ &            0.6492 &            0.6509 &            0.6489 &            0.6497 &            0.6509 &            0.0009 \\
              & $\lambda=10$ &            0.6507 &            0.6538 &            0.6508 &            0.6518 &            0.6538 & 0.0014 \\
        \bottomrule
        \end{tabular}
\end{table*}

\newcommand{\plotmultimnistmultiforward}[2]{
    \begin{subfigure}[b]{\qwidth\textwidth}
        \centering
        \includegraphics[width=\textwidth]{media/experiments/multimnist/multiforward/w=#1_lambda=#2_beta=0.pdf}
        \caption{\( W=#1 \) and \( \lambda=#2 \)}
    \end{subfigure}
}

\newcommand{\plotmultimnistmultiforwardval}[2]{
    \begin{subfigure}[b]{\qwidth\textwidth}
        \centering
        \includegraphics[width=\textwidth]{media/experiments/multimnist/multiforward2/v2_w=#1_lambda=#2_beta=0_val.pdf}
        \caption{\( W=#1 \) and \( \lambda=#2 \)}
    \end{subfigure}
}
\def\separation{10pt}

\begin{figure*}[h]
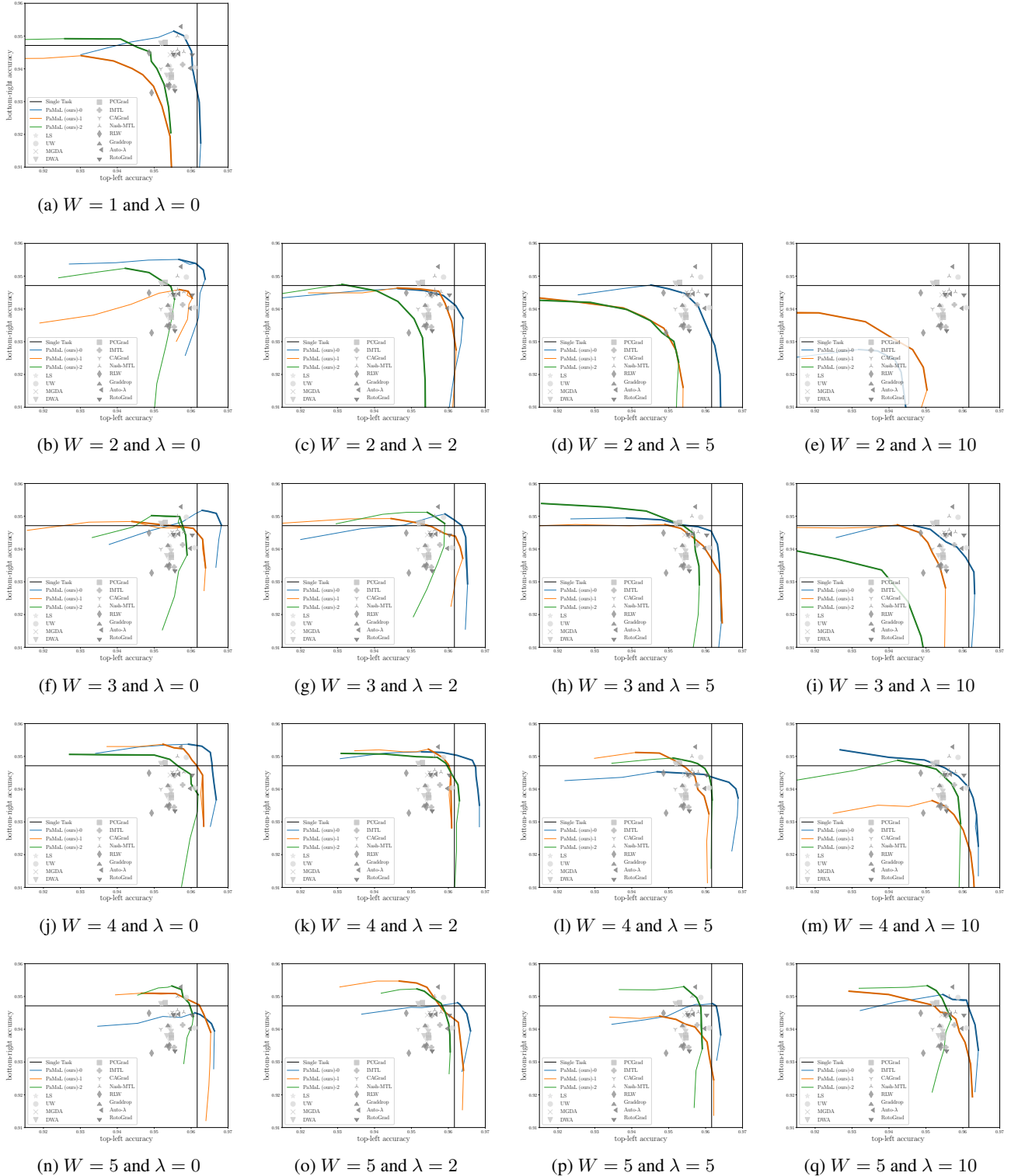

    \centering
    \def\qwidth{0.23}
    \plotmultimnistmultiforwardval{1}{0} \hfill
    \hspace*{0.5\textwidth}
    \\[\separation]

    \def\wwww{2}
    \plotmultimnistmultiforwardval{\wwww}{0} \hfill
    \plotmultimnistmultiforwardval{\wwww}{2} \hfill
    \plotmultimnistmultiforwardval{\wwww}{5} \hfill
    \plotmultimnistmultiforwardval{\wwww}{10} 
    \\[\separation]

    \def\wwww{3}
    \plotmultimnistmultiforwardval{\wwww}{0} \hfill
    \plotmultimnistmultiforwardval{\wwww}{2} \hfill
    \plotmultimnistmultiforwardval{\wwww}{5} \hfill
    \plotmultimnistmultiforwardval{\wwww}{10} 
    \\[\separation]

    \def\wwww{4}
    \plotmultimnistmultiforwardval{\wwww}{0} \hfill
    \plotmultimnistmultiforwardval{\wwww}{2} \hfill
    \plotmultimnistmultiforwardval{\wwww}{5} \hfill
    \plotmultimnistmultiforwardval{\wwww}{10} 
    \\[\separation]
   
    \def\wwww{5}
    \plotmultimnistmultiforwardval{\wwww}{0} \hfill
    \plotmultimnistmultiforwardval{\wwww}{2} \hfill
    \plotmultimnistmultiforwardval{\wwww}{5} \hfill
    \plotmultimnistmultiforwardval{\wwww}{10} \\
    
    \caption{\multimnist: Effect of multi-forward on the window \( W \) and the regularization coefficient \( \lambda \) on the \textit{validation} dataset.  The case of no multi-forward (\( W=1 \)) is presented in the first row. Multi-forward regularization for higher \( W \) values is beneficial. Intuitively, attaching serious weight on the regularization \( \lambda \in\{5, 10\} \) while sampling few times \( W\in\{2,3\} \) leads to suboptimal performance since the update step focuses on an uninformed regularization term. The accompanying quantitative analysis appears in \Autoref{tab:hv analysis:multimnist}.}
    \label{fig:multimnist-multiforward}
\end{figure*}

\newcommand{\plotcensusmultiforwardval}[2]{
    \begin{subfigure}[b]{\qwidth\textwidth}
        \centering
        \includegraphics[width=\textwidth]{media/experiments/census/multiforward2/v2_enc=256_num=#1_regcoeff=#2_beta=0_val.pdf}
        \caption{\( W=#1 \) and \( \lambda=#2 \)}
    \end{subfigure}
}

\begin{figure*}[h]
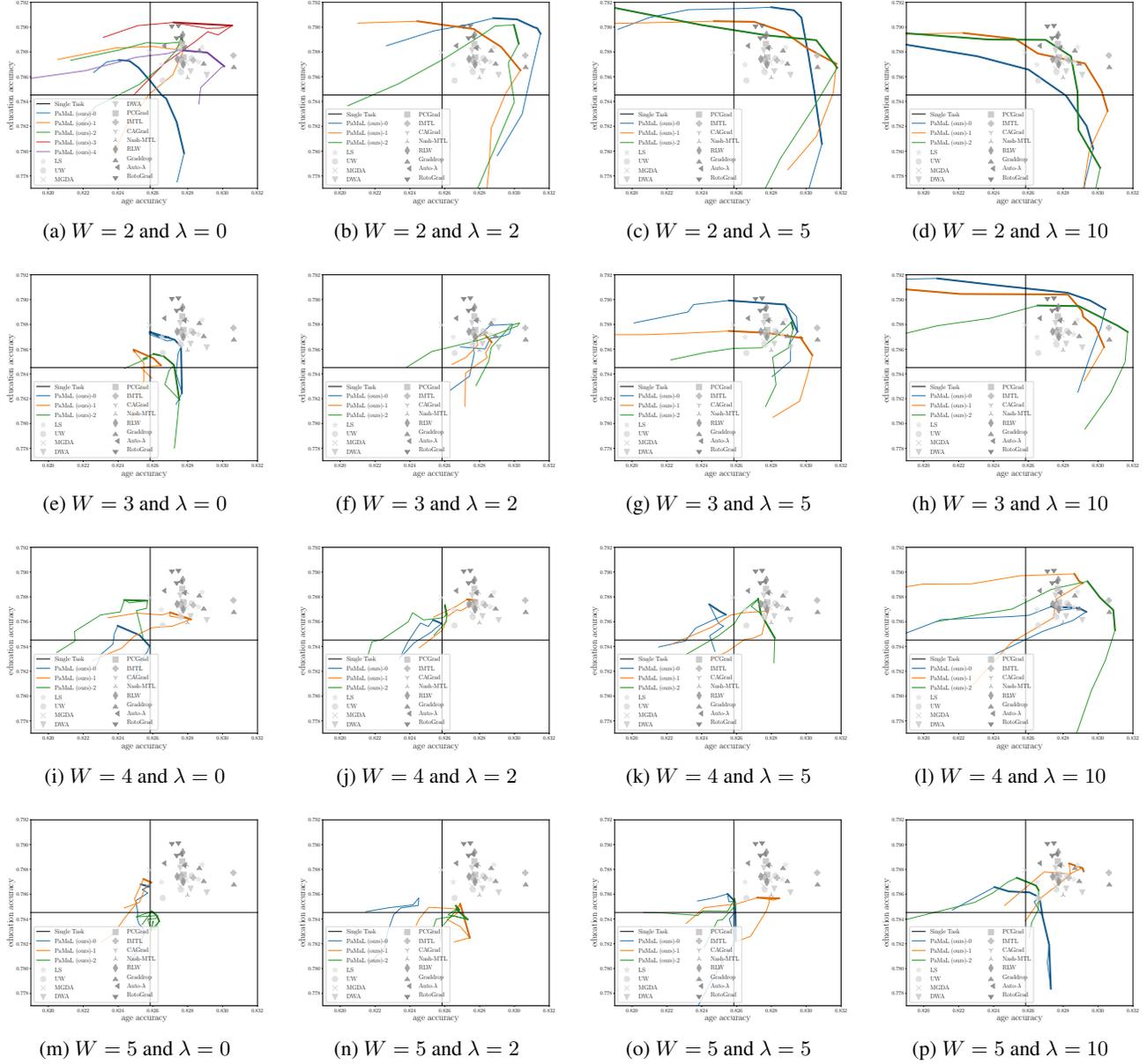

    \centering
    \def\qwidth{0.23}
    \def\separation{10pt}
    \def\wwww{2}
    \plotcensusmultiforwardval{\wwww}{0} \hfill
    \plotcensusmultiforwardval{\wwww}{2} \hfill
    \plotcensusmultiforwardval{\wwww}{5} \hfill
    \plotcensusmultiforwardval{\wwww}{10} 
    \\[\separation]

    \def\wwww{3}
    \plotcensusmultiforwardval{\wwww}{0} \hfill
    \plotcensusmultiforwardval{\wwww}{2} \hfill
    \plotcensusmultiforwardval{\wwww}{5} \hfill
    \plotcensusmultiforwardval{\wwww}{10} 
    \\[\separation]

    \def\wwww{4}
    \plotcensusmultiforwardval{\wwww}{0} \hfill
    \plotcensusmultiforwardval{\wwww}{2} \hfill
    \plotcensusmultiforwardval{\wwww}{5} \hfill
    \plotcensusmultiforwardval{\wwww}{10} 
    \\[\separation]
   
    \def\wwww{5}
    \plotcensusmultiforwardval{\wwww}{0} \hfill
    \plotcensusmultiforwardval{\wwww}{2} \hfill
    \plotcensusmultiforwardval{\wwww}{5} \hfill
    \plotcensusmultiforwardval{\wwww}{10} \\
    \caption{\census: Effect of multiforward on the window \( W \) and the regularization coefficient \( \lambda \). The axes are shared across plots. Compared to \multimnist, applying multiforward on the \textit{asymmetric} \census dataset can improve accuracies and help significantly outperform the baselines. However, widening the window \( W \) (e.g., last row for \( W=5 \)) can be hindering, since larger regularization coefficients are needed. The accompanying quantitative analysis appears in \Autoref{tab:hv analysis:census}.}
    \label{fig:census-multiforward}
\end{figure*}
        \conditionalclearpage
        \subsection{Illustrative example: ablation on loss/gradient balancing schemes}  
\label{sec:illustrative details}

\begin{figure*}[!t]
    \centering
    \includegraphics[width=\textwidth]{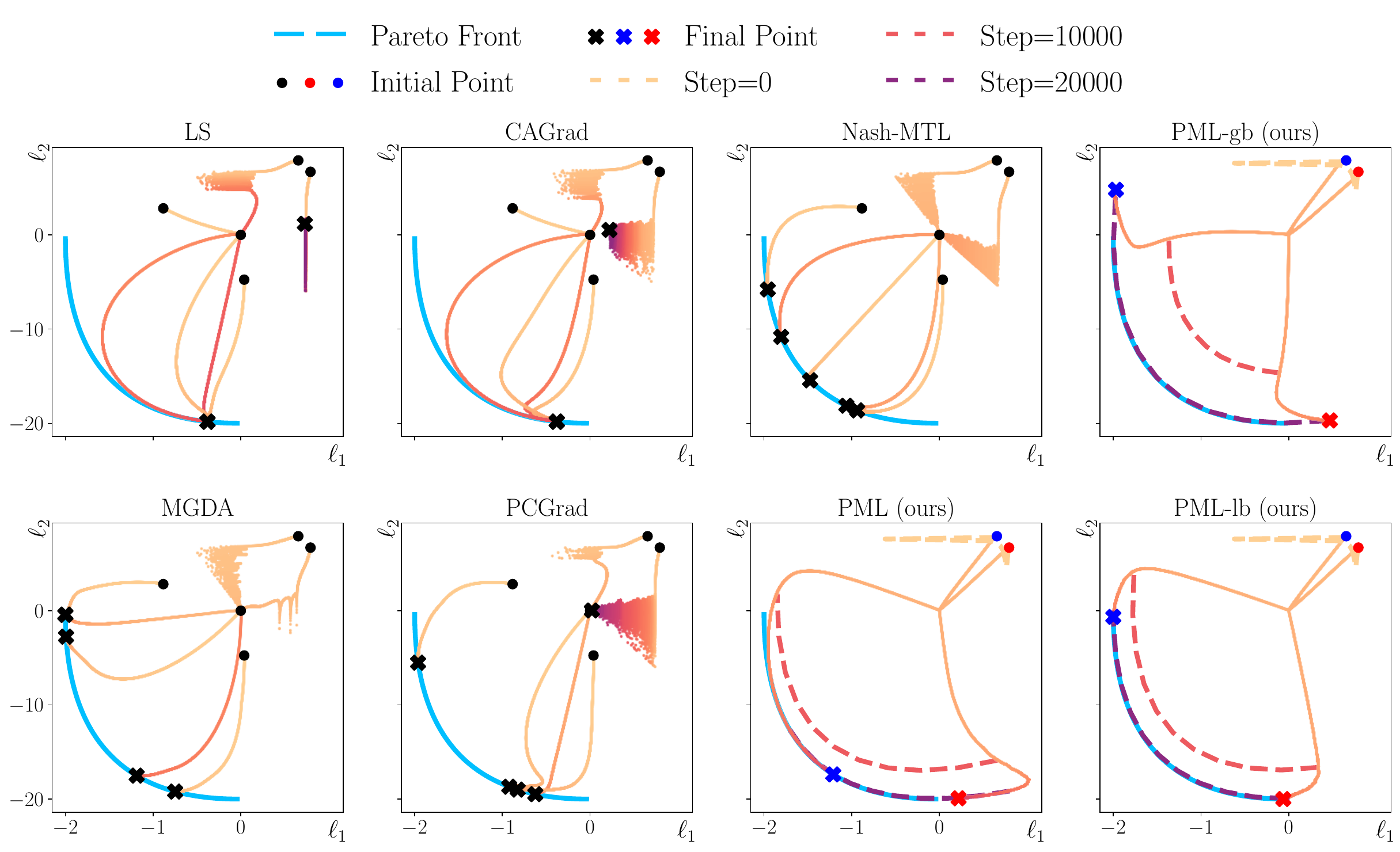}
    \caption{Optimization trajectories in objective space in the case different loss scales. Similar to \Autoref{fig:illustrative}, 5 initializations are shown for baselines and a pair of initializations for \pmlfull(\pml), in color for clarity. Dashed lines show the evolution of the mapping in loss space for the subspace at the current step. We also show the initial subspace (step\( =0 \)).
    All baselines, except Nash-MTL, and MGDA to a lesser degree, are characterized by trajectories focused on a subset of the Pareto Front, namely minimizing the task with high loss magnitude. The same observation applies to naïvely applying the proposed algorithm \pml, because using the same weighting for both the interpolation \textit{and} the losses attaches too much importance on the task with large loss magnitude. However, simple balancing schemes palliate this issue; gradient balancing (\pml-gb) discovers a superset of the Pareto Front and loss balancing (\pml-lb) discovers the exact Pareto Front.}
    \label{fig:toy-comparison2-c=0.1}
\end{figure*}

The details of the illustrative example are provided in this section. We use the configuration presented by \citet{Navon_Shamsian_Achituve_etal_2022}, which was introduced with slight modifications by \citet{Liu_Liu_Jin_etal_2021} and \citet{Yu_Kumar_Gupta_etal_2020}. Specifically, let $\bmth=\left(\theta_{1}, \theta_{2}\right) \in \mathbb{R}^{2}$ be the parameter vector and \( \vectorloss=(\tilde{\ell}_{1}, \tilde{\ell}_{2}) \) be the vector objective defined as follows:

\begin{equation*}
\tilde{\ell}_{1}(\bmth)=c_{1}(\bmth) f_{1}(\bmth)+c_{2}(\bmth) g_{1}(\bmth) \quad \text { and } \quad \tilde{\ell}_{2}(\bmth)=c_{1}(\bmth) f_{2}(\bmth)+c_{2}(\bmth) g_{2}(\bmth)
\end{equation*}
where
\begin{align*}
&f_{1}(\bmth)=\log \left(\max \left(\left|0.5\left(-\theta_{1}-7\right)-\tanh \left(-\theta_{2}\right)\right|, 5 e-6\right)\right)+6, \\
&f_{2}(\bmth)=\log \left(\max \left(\left|0.5\left(-\theta_{1}+3\right)-\tanh \left(-\theta_{2}\right)+2\right|, 5 e-6\right)\right)+6, \\
&g_{1}(\bmth)=\left(\left(-\theta_{1}+7\right)^{2}+0.1 \cdot\left(-\theta_{2}-8\right)^{2}\right) / 10-20, \\
&g_{2}(\bmth)=\left(\left(-\theta_{1}-7\right)^{2}+0.1 \cdot\left(-\theta_{2}-8\right)^{2}\right) / 10-20, \\
&c_{1}(\bmth)=\max \left(\tanh \left(0.5 \theta_{2}\right), 0\right) \quad \text { and } \quad c_{2}(\bmth)=\max \left(\tanh \left(-0.5 \theta_{2}\right), 0\right)
\end{align*}

We use the experimental setting outlined by \citet{Navon_Shamsian_Achituve_etal_2022} with minor modifications, i.e., Adam optimizer with a learning rate of \( 2e-3 \) and training lasts for \( 50K \) iterations. The overall objectives are  $\ell_{1}=c \cdot \tilde{\ell}_{1}$ and $\ell_{2}=\tilde{\ell}_{2}$ 
where we explore two configurations for the scalar \( c \), namely \( c\in\{0.1, 1\} \). For \( c=1 \), the two tasks have losses at the same scale. For \( c=0.1 \), the difference in loss scales makes the problem more challenging and the algorithm used should be characterized by scale invariance in order to find diverse solutions spanning the entirety of the Pareto Front.  The initialization points are drawn from the following set $\{(-8.5,7.5),(0.0,0.0),(9.0,9.0),(-7.5,-0.5),(9,-1.0)\}$. In the case of \pmlfull with two ensemble members there are \( 5^2=25 \) initialization pairs. In the main text we use the initialization pair with the worst initial objective values.

\Autoref{fig:toy-comparison2-c=0.1} presents the results for the case of different loss scales, i.e.,  \( c=0.1 \). We plot various baselines and three versions of the proposed algorithm, \pmlfull or \pml in short. We focus on the effect of the balancing schemes, introduced in \Autoref{sec:algo extensions}, resulting in the use of no balancing scheme (denoted as \pml), the use of gradient balancing (denoted as \pml-gb) and the use of loss balancing (denoted as \pml-lb).
We dedicate two figures for each version of the algorithm and we present all 25 initialization pairs for completeness. \Autoref{fig:toy-all-1.0} corresponds to no balancing scheme in the case of equal loss scales \( c=1.0 \), i.e., they complement \Autoref{fig:illustrative} of the main text. The subsequent figures focus on the case of unequal loss scales where \( c=0.1 \); \Autoref{fig:toy-all-0.1} corresponds to no balancing scheme, \Autoref{fig:pmlgradnorm:toy-all-0.1} corresponds to the use of gradient balancing, \Autoref{fig:rwa:toy-all-0.1} corresponds to the use of loss balancing. The first figures of each pair show the trajectories for each initialization pair, with markers for initial and final positions. The other figures of each pair dispense of the visual clutter and focus on the subspace discovered in the final step of training, which is plotted with dashed lines along with the analytical Pareto Front in solid light blue. Hence, they provide a succinct overview of whether the method was able or not to discover the (entire) Pareto Front.

For \( c=1.0 \), the proposed method is able to retrieve the exact Pareto Front with no balancing scheme for most initialization pairs. In three cases (out of 25), the method fails. In our experiments, we found that allowing longer training times or higher learning rates resolve the remaining cases.
For \( c=0.1 \), the problem is more challenging and the vanilla version of the algorithm results in a subset of the analytical Pareto Front. This subset is consistent across initialization pairs, excluding the ones the method fails, and focuses on the task with higher loss magnitude. Applying gradient balancing, shown in \Autoref{fig:pmlgradnorm:toy-all-0.1}, allows the method to retrieve (a superset of) the Pareto Front for all initialization pairs. Similarly, loss balancing, shown in \Autoref{fig:rwa:toy-all-0.1}, results in the exact Pareto Front. Hence, the inclusion of balancing schemes endows scale invariance in the proposed algorithm. Balancing schemes are used for the more challenging datasets, such as \cityscapes.

\def\version{v3}

\begin{figure*}[htpb]
    \centering
    \includegraphics[width=\linewidth]{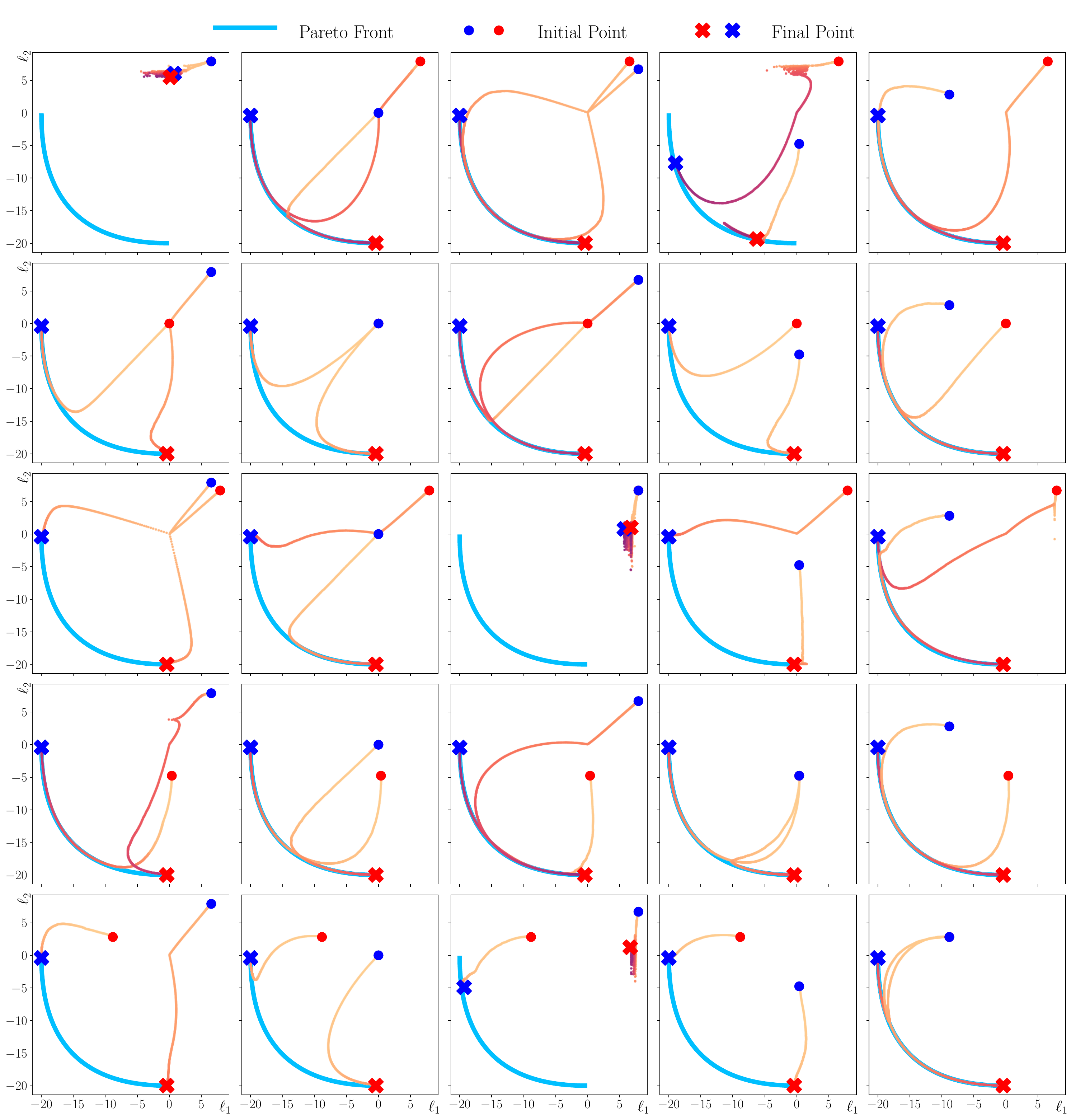}
    \caption{\textit{Illustrative example}. Optimization trajectories in objective space for all initialization pairs in the case of equal loss scales \( (c=1.0) \) and application of the proposed method with no balancing scheme. Blue and red markers show each ensemble member's loss value, dots and ``X"s correspond to the initial and final step, accordingly. In all but four cases, \pmlfull retrieves the entirety of the Pareto Front. Allowing longer training times or higher learning rates solves the remaining initialization pairs.}
    \label{fig:toy-all-1.0}
\end{figure*}

\begin{figure*}[htpb]
    \centering
    \includegraphics[width=\linewidth]{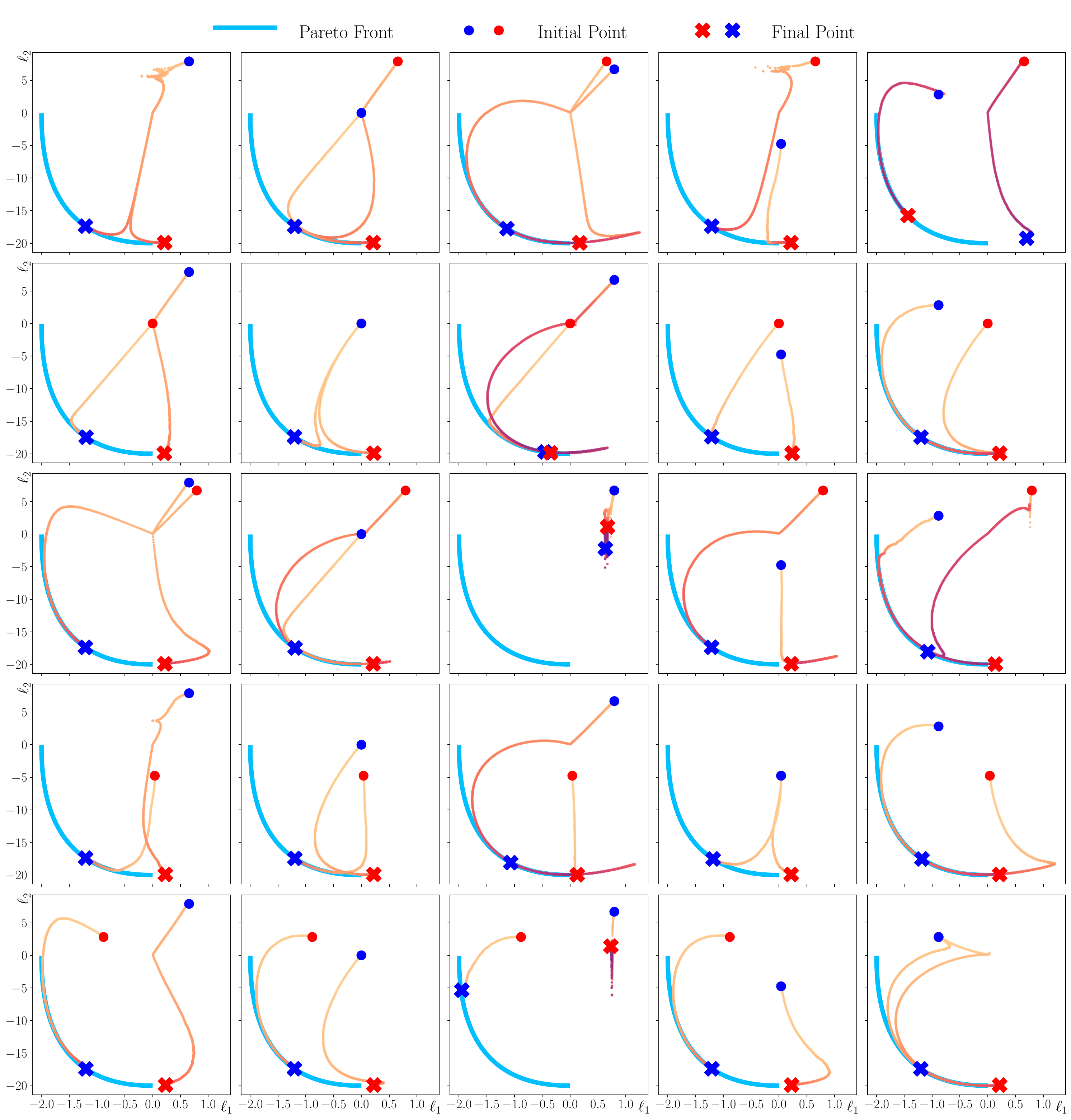}
    \caption{\textit{Illustrative example}. Optimization trajectories in objective space for all initialization pairs in the case of unequal loss scales \( (c=0.1) \) and application of the proposed method with no balancing scheme. Blue and red markers show each ensemble member's loss value, dots and ``X"s correspond to the initial and final step, accordingly. For the vast majority of initialization pairs, the lack of balancing scheme guides the ensemble to a subset of the Pareto Front, influenced by the task with higher loss magnitude. }
    \label{fig:toy-all-0.1}
\end{figure*}

\begin{figure*}[htpb]
    \centering
    \includegraphics[width=\linewidth]{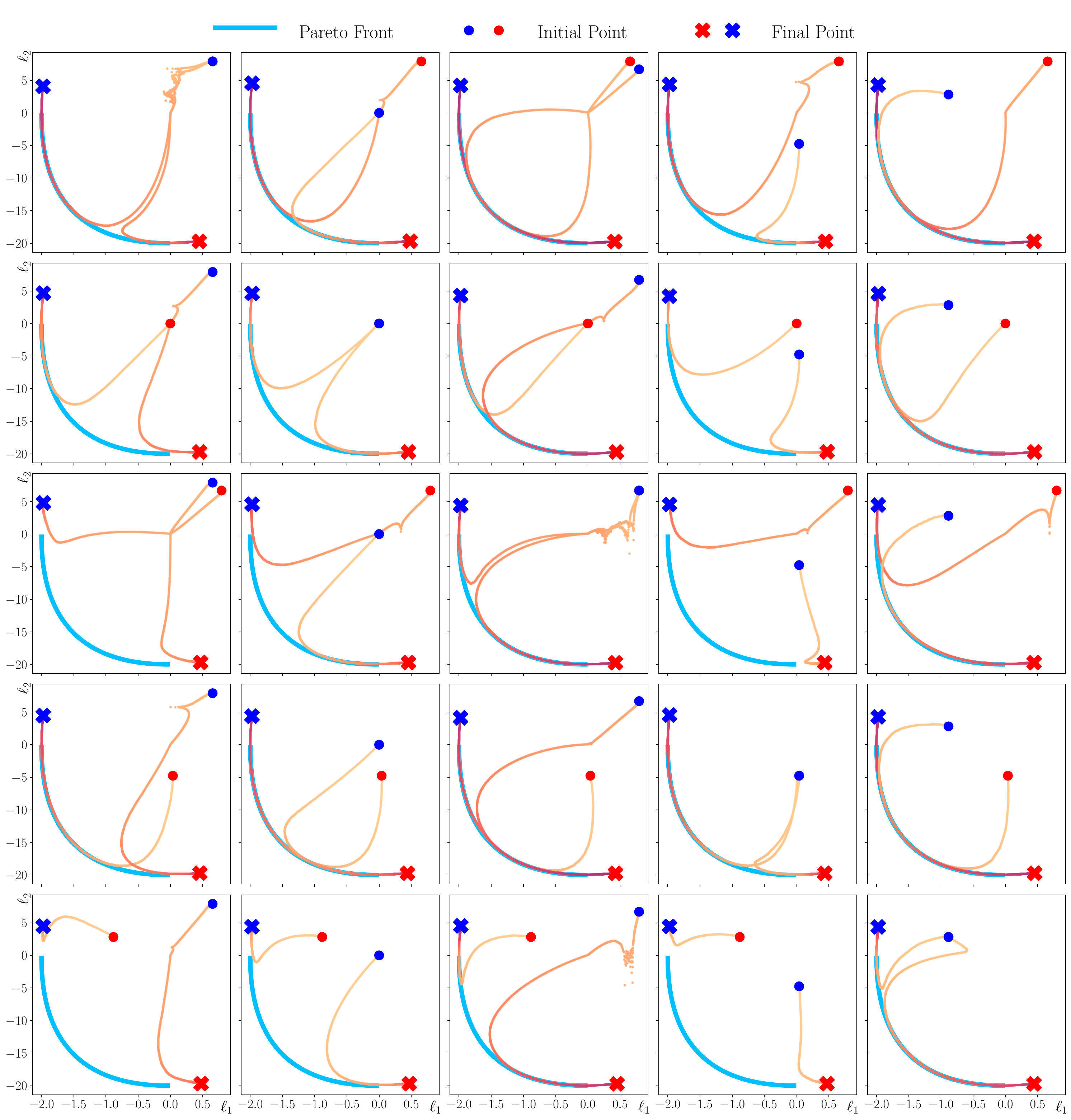}
    \caption{\textit{Illustrative example}. Optimization trajectories in objective space for all initialization pairs in the case of unequal loss scales \( (c=0.1) \) and application of the proposed method with gradient balancing scheme. Blue and red markers show each ensemble member's loss value, dots and ``X"s correspond to the initial and final step, accordingly. The proposed method discovers a subspace whose mapping in objective space results in a superset of the Pareto Front.}
    \label{fig:pmlgradnorm:toy-all-0.1}
\end{figure*}

\begin{figure*}[htpb]
    \centering
    \includegraphics[width=\linewidth]{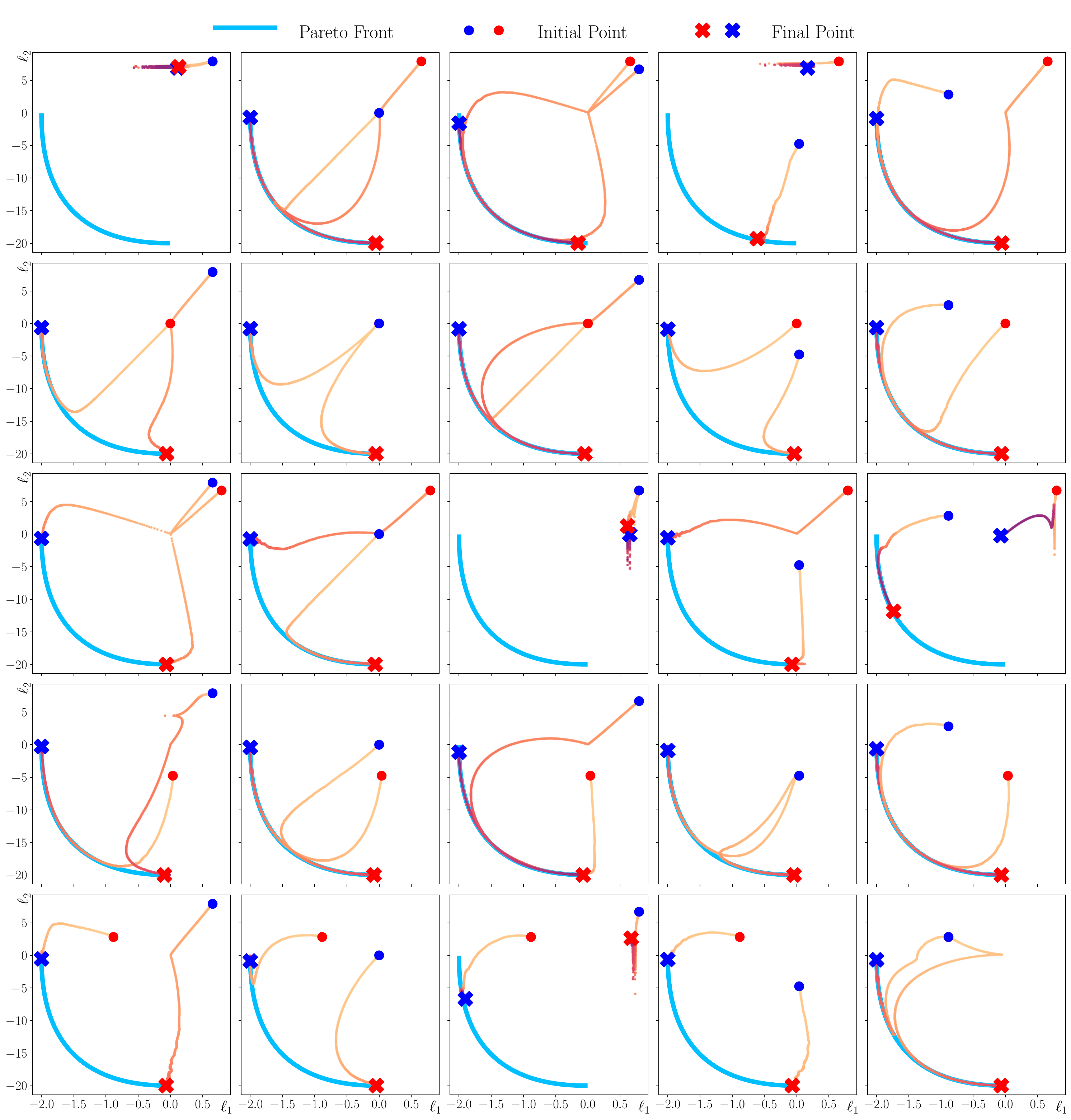}
    \caption{\textit{Illustrative example}. Optimization trajectories in objective space for all initialization pairs in the case of unequal loss scales \( (c=0.1) \) and application of the proposed method with loss balancing scheme. Blue and red markers show each ensemble member's loss value, dots and ``X"s correspond to the initial and final step, accordingly. For all but five cases, the proposed method discovers a subspace whose mapping in objective space results in the exact Pareto Front.}
    \label{fig:rwa:toy-all-0.1}
\end{figure*}

        \conditionalclearpage
        \subsection{\utkface: ablation on the effect of loss/gradient balancing schemes}  
\label{sec:utkface:extra experiments}

This section serves as supplementary to \Autoref{sec:beyond}.  \Autoref{tab:utkface:baselines} compares the performance of the baselines and the proposed method. We experiment without balancing schemes and with gradient-balancing, and present the results in \Autoref{fig:utkface-gn}. Together with the quantitative results, we observe that for datasets with varying task difficulties, scales, etc. the lack of balancing can be impeding. On the other hand, its inclusion makes the subspace functionally diverse and boosts overall performance. For instance, Huber loss on the task of age prediction is significantly improved.

\begin{table*}[t]
    \centering
    \small
    \caption{\utkface: Mean Accuracy and standard deviation of accuracy  (over 3 random seeds). For the proposed method (\pml), we report the mean and standard deviation of the best performance from the interpolated models in the sampled subspace. No multi-forward training is applied. We present \pmlfull with no balancing scheme, with gradient balancing (\textit{g}) and loss balancing (\textit{l}).}
    \label{tab:utkface:baselines}
    \begin{tabular}{lccc}
        \toprule
        {} &   Age $\downarrow$ & Gender $\uparrow$ & Ethnicity $\uparrow$ \\
        \midrule
        STL      &  $0.091 \pm 0.001$ &  $89.80 \pm 0.38$ &     $81.23 \pm 0.22$ \\
        \midrule
        LS             &  $0.100 \pm 0.008$ &   $90.43 \pm 0.76$ &     $80.49 \pm 1.64$ \\
        UW             &  $0.092 \pm 0.003$ &   $\bf 91.39 \pm 0.08$ &     $\bf 81.63 \pm 0.22$ \\
        MGDA           &  $0.091 \pm 0.006$ &   $90.71 \pm 0.22$ &     $77.29 \pm 0.44$ \\
        PCGrad         &  $0.102 \pm 0.008$ &   $90.36 \pm 1.56$ &     $79.96 \pm 2.94$ \\
        IMTL           &  $0.110 \pm 0.029$ &   $91.16 \pm 0.19$ &     $80.47 \pm 0.96$ \\
        Graddrop       &  $0.140 \pm 0.059$ &   $89.43 \pm 2.59$ &     $77.59 \pm 5.75$ \\
        CAGrad         &  $0.089 \pm 0.001$ &   $90.84 \pm 0.38$ &     $81.28 \pm 0.53$ \\
        RLW            &  $0.097 \pm 0.002$ &   $90.81 \pm 0.12$ &     $81.50 \pm 0.19$ \\
        Nash-MTL       &  $0.106 \pm 0.019$ &   $90.36 \pm 0.60$ &     $78.98 \pm 2.14$ \\
        Auto-$\lambda$ &  $0.091 \pm 0.003$ &   $90.84 \pm 0.35$ &     $81.58 \pm 0.06$ \\
        \midrule
        COSMOS &  $0.107 \pm 0.003$ &  $89.68 \pm 0.40$ &  $79.39 \pm 0.59$ \\
        PHN &  $0.106 \pm 0.001$ &  $90.49 \pm 0.34$ &  $79.99 \pm 0.23$ \\
        \midrule
        PAMAL-\textit{g}(W=1, p=1) &  $0.094 \pm 0.001$ &  $90.65 \pm 0.20$ &  $80.03 \pm 0.27$ \\
        PAMAL-\textit{l}(W=3, p=2) &  $0.099 \pm 0.003$ &  $90.62 \pm 0.27$ &  $80.82 \pm 0.97$ \\
        PAMAL-\textit{g}(W=3, p=2) &  $\bf 0.083 \pm 0.001$ &  $90.93 \pm 0.25$ &  $80.78 \pm 0.29$ \\
        \bottomrule
    \end{tabular}
\end{table*}
\vspace*{-10pt}
\looseness=-1

\def\qwidth{0.9}
\def\separation{1pt}
\begin{figure*}[p]
    \centering
    \begin{subfigure}[b]{\qwidth\textwidth}
        \centering
        \includegraphics[width=\textwidth]{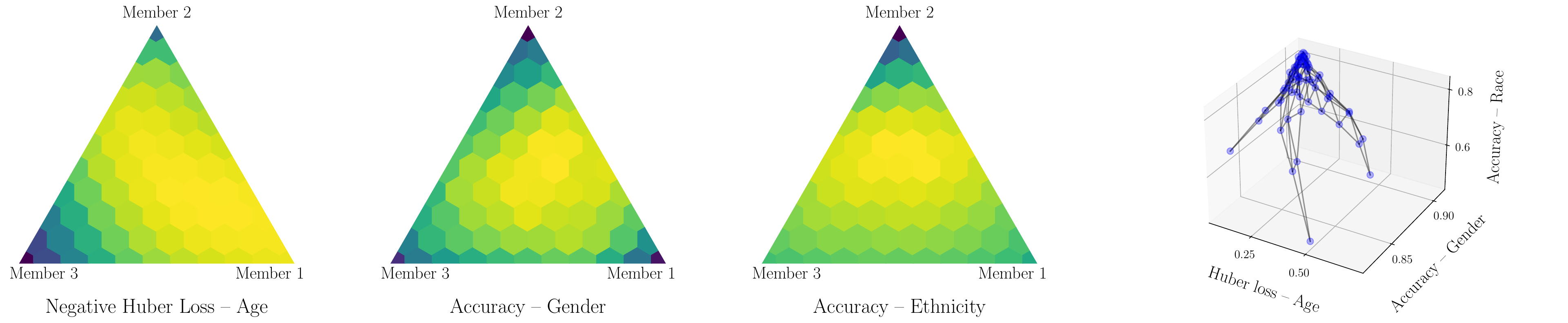}
    \end{subfigure}
    \\[\separation]
    \begin{subfigure}[b]{\qwidth\textwidth}
        \centering
        \includegraphics[width=\textwidth]{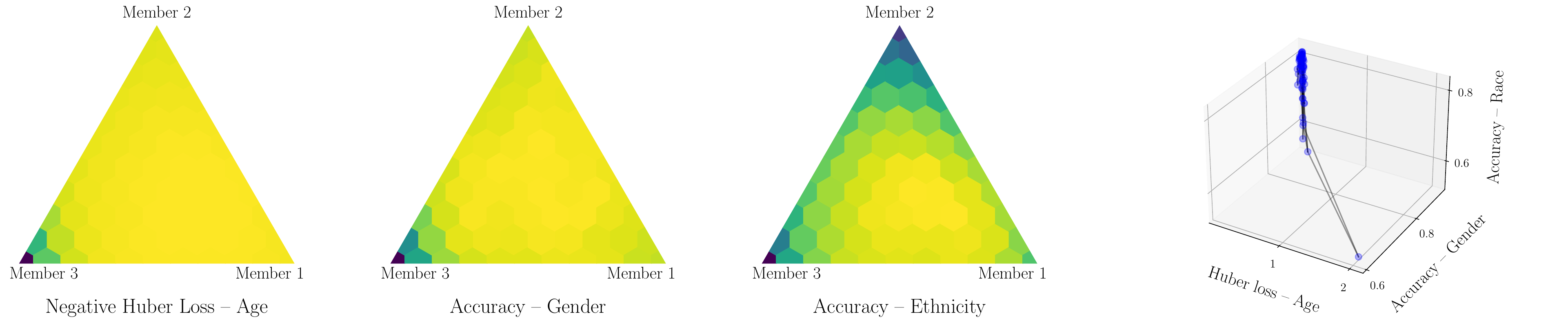}
    \end{subfigure}
    \\[\separation]
    \begin{subfigure}[b]{\qwidth\textwidth}
        \centering
        \includegraphics[width=\textwidth]{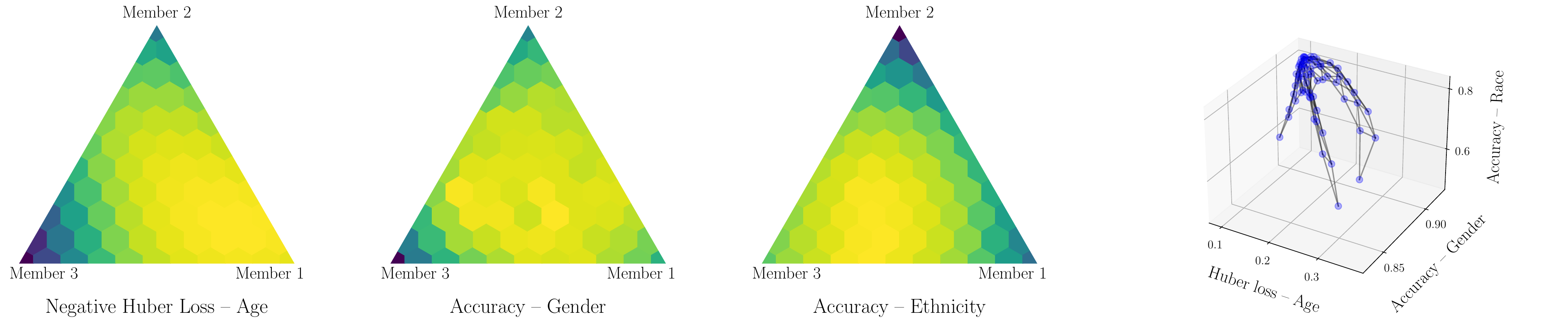}
        \caption{Linear Scalarization}
    \end{subfigure}
    \\[10pt]
    \begin{subfigure}[b]{\qwidth\textwidth}
        \centering
        \includegraphics[width=\textwidth]{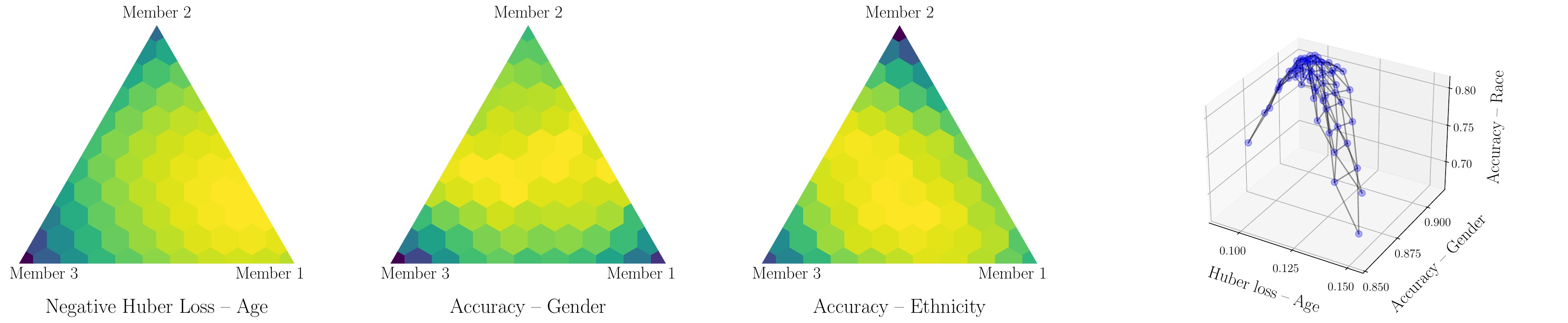}
    \end{subfigure}
    \\[\separation]
    \begin{subfigure}[b]{\qwidth\textwidth}
        \centering
        \includegraphics[width=\textwidth]{media/experiments/utkface/v3-gradnorm-seed=1.pdf}
    \end{subfigure}
    \\[\separation]
    \begin{subfigure}[b]{\qwidth\textwidth}
        \centering
        \includegraphics[width=\textwidth]{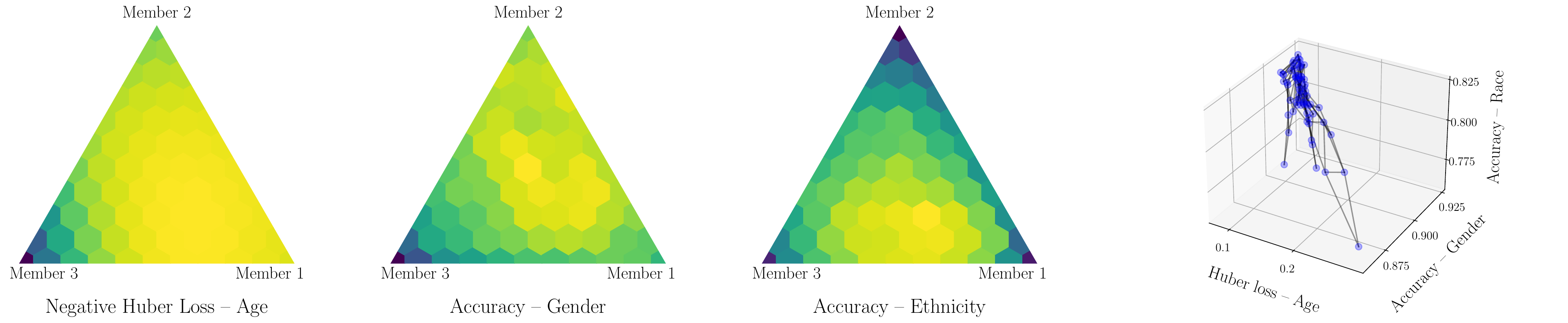}
        \caption{Gradient Balancing}
    \end{subfigure}
    \caption{\utkface results with Gradient-Balancing Scheme for all three seeds.  
    Each triangle shows the 66 points in the convex hull and color is used for the  performance on the associated task. The 3d plot shows the mapping of the subspace to the multi-objective space. For datasets with tasks of varying loss scales, applying gradient balancing improves functional diversity and performance, as shown in \Autoref{tab:utkface:baselines}.}
    \label{fig:utkface-gn}
\end{figure*}
        \subsection{Hyperparameter optimization for PHN and COSMOS}
\label{sec:appendix:phn and cosmos ablation}

\paragraph*{PHN - Pareto HyperNetwork \citep{Navon_Shamsian_Fetaya_etal_2021}} The method has one hyperparameter: \( p \) for the sampling of the Dirichlet distribution and two solvers: Linear and EPO \citep{Mahapatra_Rajan_2020}. The method also requires the definition of the architecture of the HyperNetwork. Following the authors' implementation we use a MLP with 2 hidden layers of \( w=100 \) dimensions each. The output layer of the MLP has as many neurons as the target network, e.g. a LeNet for \multimnist experiments or a ResNet18 for \utkface. Essentially, this means that the HyperNetwork requires at least \( w \) times the number of the target network parameters. The models for \utkface and \cityscapes have approximately 11M and 17M parameters, leading to the HyperNetwork to
have 1.1B and 1.7B parameters. Hence, due to computational reasons, the PHN baseline for these benchmarks applies chunking \citep{Ha_Dai_Le_2017}. For \cityscapes we were unable to retrieve good results for PHN using chunking and, hence we omit it from the main text and from additional experiments in the appendix.

For the concentration parameter \( p \) we ablate over the values \( \{0.001, 0.1, 0.2(\text{used in the original paper}), 0.5, 1, 2\} \) and over the two solvers, i.e., Linear and EPO. For \multimnist, the original paper used 100 epochs and complex schedulers with a lower learning rate of \( 10^{-4} \). We limit the training budget to 10 epochs for all baselines and omit scheduling for \multimnist. However, we also consider the lower learning rate in the ablation. The full hyperparameter sweep is presented in \autoref{table:appendix:ablation:phn - multimnist} for \multimnist and \autoref{table:appendix:ablation:phn - census} for \census.

For each hyperparameter configuration, three seeds are considered and the best configuration is selected by the best average (across seeds) HyperVolume score that also satisfies the criterion that the average (across seeds) Spearman correlation over the two task performances is lower than a threshold. Consider the case of two classification tasks, where performance is gauged by validation accuracy. Then, a solution is examined over the different tradeoffs \( \calA=\{{(\alpha, 1-\alpha): \alpha\in [0,1]}\} \), and, performance of one task should monotonically increase as \( \alpha \) is increased and vice versa for the other task. In this optimal scenario, the Spearman correlation would be \( -1 \). The reason for this additional criterion is that the optimal hyperparameter configurations produced degenerate solutions, as explained visually in \autoref{fig:ablation for phn and cosmos}. 
The aforementioned ablation applies to \multimnist and \census.

\paragraph*{COSMOS - Conditioned One-shot Multi-Objective Search \citep{Ruchte_Grabocka_2021}} The method has two hyperparameters: \( p \) for the sampling of the Dirichlet distribution and \( \lambda \) as the coefficient for the proposed regularization. We use the PHN search space for \( \alpha \), consider \(  \lambda\in\{0.1, 1, 2, 5, 8\} \) and learning rate \( \in \{10^{-3}, 10^{-4}\} \). The full hyperparameter sweep is presented in \autoref{table:appendix:ablation:cosmos - multimnist} for \multimnist and \autoref{table:appendix:ablation:cosmos - census} for \census.

\begin{figure}[t]
    \centering
    \includegraphics[width=0.95\textwidth]{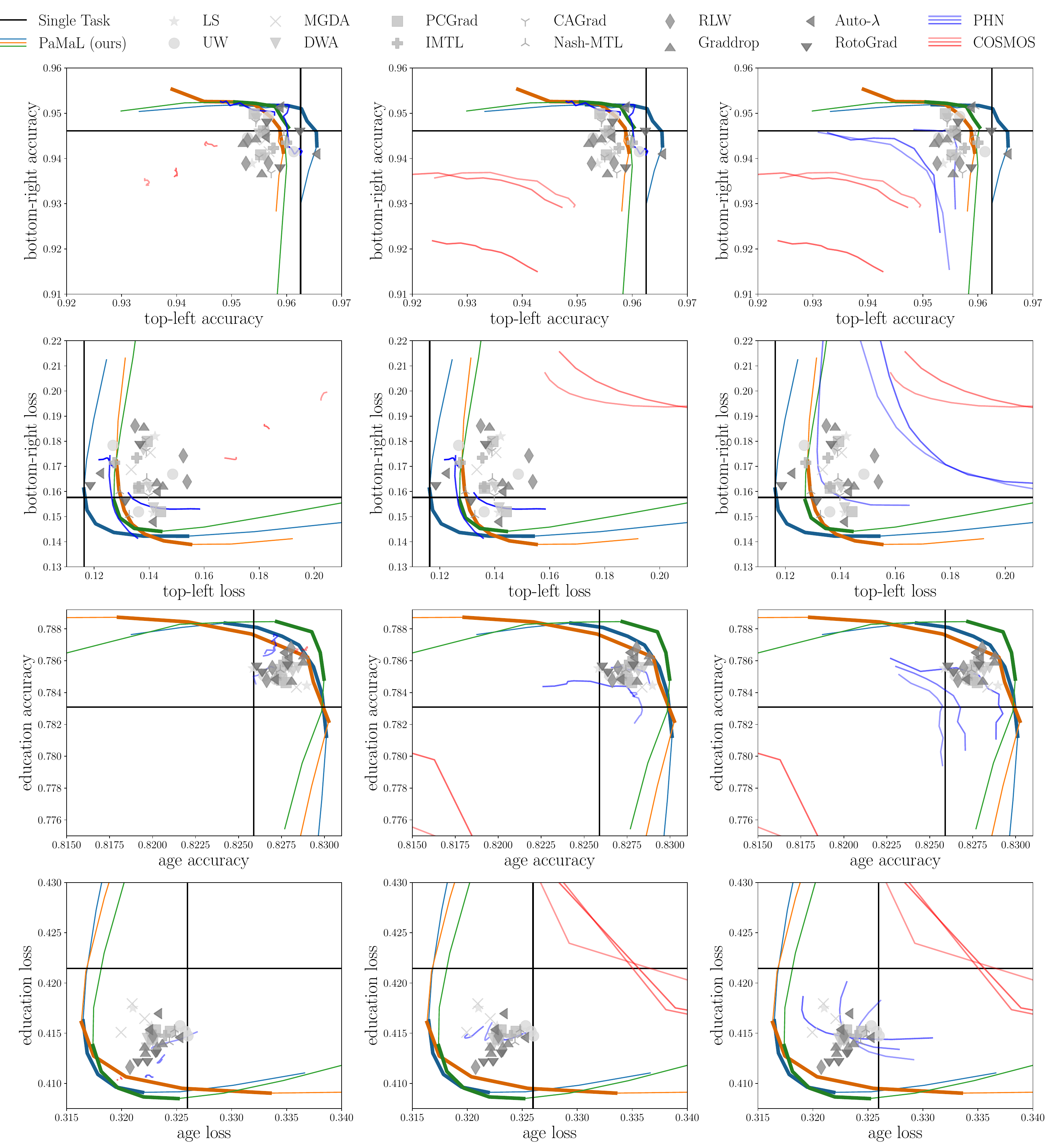}
    \caption{Effect of the Spearman correlation (S) as a secondary criterion on the PHN and COSMOS baselines. The other methods are fixed per row. Zoom for details. (Left) no additional constraint, (Middle) \( S< -0.5 \), (Right) \( S<-0.6 \). (First row) \multimnist accuracy, (Second row) \multimnist loss, (Third row) \census accuracy, (Fourth row) \census loss.  The exclusion of the Spearman threshold criterion leads to degenerate solutions (left column) that are not in the spirit of the original method; either a single point lies in the Pareto Front or the mapping from desired trade-off to network weights is obfuscated. Additionally, loss curves are generally "smoother" than accuracy curves, attesting to the generalization gap.}
    \label{fig:ablation for phn and cosmos}
\end{figure}

\begin{table}[t!]
    \centering
    \small
    \caption{Hyperparameter search for COSMOS on \multimnist. Results refer to the validation set. Higher HyperVolume is better. Lower Spearman correlation is better. The accuracy columns refer to the maximum accuracy sampled by conditioning the network with ten different user preferences.}
    \label{table:appendix:ablation:cosmos - multimnist}
    \begin{tabular}{lllrrrr}
\toprule
$\alpha$ & $\lambda$ & lr  &  acc top-left &  acc bottom-right &      HV &  Spearman \\
\midrule
\multirow{4}{*}{0.1} & 1.0 & 0.001 &        0.9440 &            0.9249 &  0.8731 &   -0.9838 \\
    & 2.0 & 0.001 &        0.9406 &            0.9219 &  0.8666 &   -0.9960 \\
    & 5.0 & 0.001 &        0.9400 &            0.9236 &  0.8667 &   -0.9919 \\
    & 8.0 & 0.001 &        0.9402 &            0.9219 &  0.8647 &   -0.9950 \\
\cmidrule{1-7}
\multirow{4}{*}{1.0} & 1.0 & 0.001 &        0.9339 &            0.9194 &  0.8587 &   -0.8121 \\
    & 2.0 & 0.001 &        0.9294 &            0.9173 &  0.8515 &   -0.8061 \\
    & 5.0 & 0.001 &        0.9279 &            0.9055 &  0.8380 &   -0.8828 \\
    & 8.0 & 0.001 &        0.9293 &            0.9059 &  0.8395 &   -0.9636 \\
\cmidrule{1-7}
\multirow{4}{*}{2.0} & 1.0 & 0.001 &        0.9422 &            0.9277 &  0.8741 &   -0.1328 \\
    & 2.0 & 0.001 &        0.9304 &            0.9179 &  0.8530 &   -0.8424 \\
    & 5.0 & 0.001 &        0.9285 &            0.9159 &  0.8487 &   -0.9434 \\
    & 8.0 & 0.001 &        0.9240 &            0.9120 &  0.8412 &   -0.9030 \\
\cmidrule{1-7}
\multirow{4}{*}{5.0} & 1.0 & 0.001 &        0.9378 &            0.9326 &  0.8746 &    0.0866 \\
    & 2.0 & 0.001 &        0.9280 &            0.9209 &  0.8544 &   -0.9279 \\
    & 5.0 & 0.001 &        0.9230 &            0.9162 &  0.8443 &   -0.8797 \\
    & 8.0 & 0.001 &        0.9242 &            0.9028 &  0.8338 &   -0.7078 \\
\bottomrule
\end{tabular}

\end{table}

\begin{table}[b!]
    \centering
    \small
    \caption{Hyperparameter search for PHN on \census. Results refer to the validation set. Higher HyperVolume is better. Lower Spearman correlation is better. The accuracy columns refer to the maximum accuracy sampled by conditioning the network with ten different user preferences.}
    \label{table:appendix:ablation:phn - census}
    \begin{tabular}{lllrrrr}
\toprule
$\alpha$ & solver & lr       &  acc age &  acc education &      HV &  Spearman \\
\midrule
\multirow{4}{*}{0.001} & \multirow{2}{*}{EPO} & 1e-4 &   0.8277 &         0.7883 &  0.6524 &    0.2294 \\
      &        & 1e-3 &   0.8250 &         0.7869 &  0.6492 &    0.9753 \\
\cmidrule{2-7}
      & \multirow{2}{*}{linear} & 1e-4 &   0.8283 &         0.7883 &  0.6530 &    0.0226 \\
      &        & 1e-3 &   0.8271 &         0.7861 &  0.6502 &    0.0000 \\
\cmidrule{1-7}
\cmidrule{2-7}
\multirow{3}{*}{0.010} & EPO & 1e-4 &   0.8264 &         0.7860 &  0.6495 &    0.1478 \\
      & \multirow{2}{*}{linear} & 1e-4 &   0.8274 &         0.7879 &  0.6519 &   -0.7723 \\
      &        & 1e-3 &   0.8268 &         0.7874 &  0.6510 &   -0.5617 \\
\cmidrule{1-7}
\cmidrule{2-7}
\multirow{3}{*}{0.100} & EPO & 1e-4 &   0.8277 &         0.7898 &  0.6537 &    0.1538 \\
      & \multirow{2}{*}{linear} & 1e-4 &   0.8272 &         0.7895 &  0.6530 &    0.2244 \\
      &        & 1e-3 &   0.8274 &         0.7883 &  0.6523 &    0.1743 \\
\cmidrule{1-7}
\cmidrule{2-7}
\multirow{2}{*}{0.200} & EPO & 1e-4 &   0.8292 &         0.7840 &  0.6501 &   -0.1337 \\
      & linear & 1e-4 &   0.8282 &         0.7833 &  0.6487 &    0.1567 \\
\cmidrule{1-7}
\multirow{3}{*}{0.500} & \multirow{2}{*}{EPO} & 1e-4 &   0.8287 &         0.7867 &  0.6520 &   -0.0834 \\
      &        & 1e-3 &   0.8274 &         0.7856 &  0.6501 &   -0.1058 \\
\cmidrule{2-7}
      & linear & 1e-4 &   0.8270 &         0.7854 &  0.6495 &    0.1374 \\
\cmidrule{1-7}
\multirow{3}{*}{1.000} & \multirow{2}{*}{EPO} & 1e-4 &   0.8291 &         0.7877 &  0.6531 &   -0.0442 \\
      &        & 1e-3 &   0.8270 &         0.7848 &  0.6490 &   -0.3510 \\
\cmidrule{2-7}
      & linear & 1e-4 &   0.8279 &         0.7870 &  0.6515 &   -0.0356 \\
\cmidrule{1-7}
\multirow{2}{*}{2.000} & EPO & 1e-4 &   0.8297 &         0.7876 &  0.6535 &   -0.5062 \\
      & linear & 1e-4 &   0.8266 &         0.7872 &  0.6507 &    0.6878 \\
\bottomrule
\end{tabular}

\end{table}

\begin{table}[t!]
    \centering
    \small
    \caption{Hyperparameter search for PHN on \multimnist. Results refer to the validation set. Higher HyperVolume is better. Lower Spearman correlation is better. The accuracy columns refer to the maximum accuracy sampled by conditioning the network with ten different user preferences.}
    \label{table:appendix:ablation:phn - multimnist}
    \begin{tabular}{lllrrrr}
\toprule
$\alpha$ & $\lambda$ & lr  &  acc top-left &  acc bottom-right &      HV &  Spearman \\
\midrule
\multirow{2}{*}{0.001} & EPO & 0.001 &        0.9570 &            0.9290 &  0.8890 &   -0.9464 \\
      & linear & 0.001 &        0.9555 &            0.9440 &  0.9019 &   -0.9838 \\
\cmidrule{1-7}
\multirow{2}{*}{0.010} & EPO & 0.001 &        0.9615 &            0.8874 &  0.8532 &   -0.2705 \\
      & linear & 0.001 &        0.9587 &            0.9381 &  0.8978 &   -1.0000 \\
\cmidrule{1-7}
\multirow{2}{*}{0.100} & EPO & 0.001 &        0.9446 &            0.9257 &  0.8743 &   -0.9636 \\
      & linear & 0.001 &        0.9547 &            0.9406 &  0.8978 &   -0.9960 \\
\cmidrule{1-7}
\multirow{2}{*}{0.200} & EPO & 0.001 &        0.9502 &            0.9411 &  0.8941 &   -0.9394 \\
      & linear & 0.001 &        0.9468 &            0.9256 &  0.8763 &   -0.9919 \\
\cmidrule{1-7}
\multirow{2}{*}{0.500} & EPO & 0.001 &        0.9552 &            0.9436 &  0.9012 &   -0.9818 \\
      & linear & 0.001 &        0.9585 &            0.9495 &  0.9101 &   -0.7858 \\
\cmidrule{1-7}
\multirow{2}{*}{1.000} & EPO & 0.001 &        0.9500 &            0.9369 &  0.8900 &   -0.8858 \\
      & linear & 0.001 &        0.9579 &            0.9429 &  0.9032 &   -0.0016 \\
\cmidrule{1-7}
\multirow{2}{*}{2.000} & EPO & 0.001 &        0.9548 &            0.9438 &  0.9011 &   -0.7765 \\
      & linear & 0.001 &        0.9570 &            0.9384 &  0.8981 &   -0.6209 \\
\bottomrule
\end{tabular}

\end{table}

\begin{table}[b!]
    \centering
    \small
    \caption{Hyperparameter search for COSMOS on \census. Results refer to the validation set. Higher HyperVolume is better. Lower Spearman correlation is better. The accuracy columns refer to the maximum accuracy sampled by conditioning the network with ten different user preferences.}
    \label{table:appendix:ablation:cosmos - census}
    \begin{tabular}{cccccccccccccccccc}
\toprule
     &     & \multicolumn{4}{c}{lr=1e-3} &  & \multicolumn{4}{c}{lr=1e-4} \\
\cmidrule{8-11}
\cmidrule{3-6}
$\alpha$ & $\lambda$ &   acc age & acc education &      HV & Spearman & &   acc age & acc education &      HV & Spearman \\
\midrule
\multirow{4}{*}{0.01} & 0.1 &   0.8284 &        0.7902 &  0.6546 &   0.3204 &         &    0.8281 &        0.7891 &  0.6535 &   0.5075 \\
     & 1.0 &   0.8258 &        0.7872 &  0.6500 &  -0.2898 &         &    0.8254 &        0.7892 &  0.6514 &   0.4072 \\
     & 2.0 &   0.8118 &        0.7874 &  0.6370 &  -0.9960 &         &    0.8190 &        0.7884 &  0.6447 &  -0.9919 \\
     & 5.0 &   0.8105 &        0.7871 &  0.6298 &  -1.0000 &         &    0.8075 &        0.7878 &  0.6301 &  -1.0000 \\
\cmidrule{1-11}
\multirow{4}{*}{0.10} & 0.1 &   0.8287 &        0.7892 &  0.6540 &  -0.0411 &         &    0.8285 &        0.7887 &  0.6534 &   0.0948 \\
     & 1.0 &   0.8257 &        0.7891 &  0.6516 &  -0.9667 &         &    0.8254 &        0.7894 &  0.6515 &   0.0287 \\
     & 2.0 &   0.8196 &        0.7883 &  0.6448 &  -0.9960 &         &    0.8181 &        0.7883 &  0.6441 &  -0.9677 \\
     & 5.0 &   0.8201 &        0.7892 &  0.6405 &  -0.9838 &         &    0.8083 &        0.7879 &  0.6314 &  -0.9960 \\
\cmidrule{1-11}
\multirow{4}{*}{0.20} & 0.1 &   0.8286 &        0.7887 &  0.6535 &  -0.0451 &         &    0.8284 &        0.7889 &  0.6536 &  -0.0185 \\
     & 1.0 &   0.8276 &        0.7886 &  0.6526 &  -0.5289 &         &    0.8267 &        0.7895 &  0.6527 &  -0.0051 \\
     & 2.0 &   0.8238 &        0.7902 &  0.6502 &  -0.9677 &         &    0.8206 &        0.7881 &  0.6461 &  -0.9475 \\
     & 5.0 &   0.8247 &        0.7907 &  0.6492 &  -0.9394 &         &    0.8137 &        0.7876 &  0.6356 &  -0.9950 \\
\cmidrule{1-11}
\multirow{4}{*}{0.50} & 0.1 &   0.8291 &        0.7880 &  0.6533 &   0.3141 &         &    0.8293 &        0.7892 &  0.6545 &   0.3629 \\
     & 1.0 &   0.8283 &        0.7877 &  0.6524 &  -0.2757 &         &    0.8267 &        0.7897 &  0.6528 &  -0.4077 \\
     & 2.0 &   0.8255 &        0.7892 &  0.6508 &  -0.9596 &         &    0.8223 &        0.7882 &  0.6476 &  -0.9717 \\
     & 5.0 &   0.8277 &        0.7900 &  0.6505 &  -0.9556 &         &    0.8180 &        0.7879 &  0.6387 &  -0.9798 \\
\cmidrule{1-11}
\multirow{4}{*}{1.00} & 0.1 &   0.8285 &        0.7891 &  0.6537 &   0.5783 &         &    0.8293 &        0.7891 &  0.6544 &   0.3761 \\
     & 1.0 &   0.8279 &        0.7878 &  0.6522 &  -0.6916 &         &    0.8272 &        0.7897 &  0.6532 &  -0.1814 \\
     & 2.0 &   0.8290 &        0.7886 &  0.6537 &  -0.3253 &         &    0.8239 &        0.7885 &  0.6496 &  -0.3459 \\
     & 5.0 &   0.8283 &        0.7898 &  0.6533 &  -0.8384 &         &    0.8218 &        0.7874 &  0.6422 &  -0.8909 \\
\cmidrule{1-11}
\multirow{4}{*}{2.00} & 0.1 &   0.8280 &        0.7888 &  0.6532 &  -0.2172 &         &    0.8294 &        0.7896 &  0.6549 &  -0.0839 \\
     & 1.0 &   0.8290 &        0.7875 &  0.6528 &  -0.8392 &         &    0.8271 &        0.7900 &  0.6534 &  -0.3673 \\
     & 2.0 &   0.8276 &        0.7891 &  0.6530 &  -0.8489 &         &    0.8248 &        0.7888 &  0.6506 &   0.3466 \\
     & 5.0 &   0.8286 &        0.7896 &  0.6531 &  -0.7423 &         &    0.8234 &        0.7870 &  0.6441 &  -0.8061 \\
\bottomrule
\end{tabular}

\end{table}

        \conditionalclearpage

        \section{Additional Experiments}
        \label{appendix:more experiments}
        \subsection{Details on experimental configurations}
\label{sec:experimental details}  

\paragraph*{\multimnist}

\multimnist is a synthetic dataset derived form the samples of \mnist. Since there is no publicly available version, we create our own by the following procedure. For each \multimnist image, we sample (with replacement) two \mnist images (of size \( 28\times 28 \)) and place them top-left and bottom-right on a \( 36\times 36 \) grid. This grid is then resized to \( 28\times 28 \) pixels. The procedure is repeated 60000 times, 10000 and 10000 times for training, validation and test datasets. 
We use a LeNet shared-bottom architecture. Specifically, the encoder has two convolutional layers with 10 and 20 channels and kernel size of 5 followed by Maxpool and a ReLU nonlinearity each. The final layer of the encoder is fully connected producing an embedding with 50 features. The decoders are fully connected with two layers, one with 50 features and the output layer has 10. We use Adam optimizer \cite{Kingma_Ba_2015} with learning rate \( 10^{-3} \), no scheduler and the batch size is set to 256. Training lasts 10 epochs.

\paragraph*{\census}

The original version of the \census\cite{Kohavi_1996}  dataset has one task: predicting whether a person's income exceeds \$50000. The dataset becomes suitable for \mtl by turning one or several features to tasks \citep{Lin_Zhen_Li_etal_2019}. We focus on the task combination of predicting age and education level, similar to \citet{Ma_Du_Matusik_2020}.
The model has a Multi-Layer Perceptron shared-bottom architecture. The encoder has one layer with \( 256 \) neurons, followed by a ReLU nonlinearity, and two decoders with 2 output neurons each (since the tasks are binary classification). 
Training lasts 10 epochs. We use Adam optimizer learning rate of \( 10^{-3} \).

\paragraph*{\multimnistThree}
The configuration of \multimnist is used. The model has three decoders. Training lasts 20 epochs.

\paragraph*{\utkface}
The \utkface dataset has more than 20,000 face images of dimensions \( 200\times 200 \) pixels and 3 color channels. The dataset has three tasks: predicting age (modeled as regression using Huber loss - similar to \cite{Ma_Du_Matusik_2020}), classifying gender and ethnicity (modeled as classification tasks using Cross-Entropy loss). Images are resized to \( 64\times 64 \) pixels, age is normalized and a 80/20 train/test split is used. 
We use a shared-bottom architecture; the encoder is a ResNet18 \citep{He_Zhang_Ren_etal_2016} model without the last fully connected layer. The decoders (task-specific layers) consist of one fully-connected layer, where the output dimensions are 1, 2 and 5 for age (modeled as regression), gender (binary classification) and ethnicity (classification with 5 classes). Training lasts 100 epochs, batch size is 256 and we use Adam optimizer with a learning rate of \( 10^{-3} \). No scheduler is used.

\paragraph*{\cityscapes}
Our experimental configuration is very similar to prior work, namely \cite{Liu_Johns_Davison_2019,Yu_Kumar_Gupta_etal_2020,Liu_Liu_Jin_etal_2021,Navon_Shamsian_Achituve_etal_2022}. All images are resized to \( 128\times256 \). The tasks used are coarse semantic segmentation and depth regression. The task of semantic segmentation has 7 classes, whereas the original has 19. We use a SegNet architecture \citep{Badrinarayanan_Kendall_Cipolla_2017} and train the model for 100 epochs with Adam optimizer \citep{Kingma_Ba_2015} of an initial learning rate \( 10^{-4} \). We employ a  scheduler that halves the learning rate after 75 epochs.

        \conditionalclearpage
        \subsection{HyperVolume analysis on \multimnist and \census} 
\label{sec:hv analysis}
\begin{figure*}[!t]
    \centering
    \includegraphics[width=0.95\textwidth]{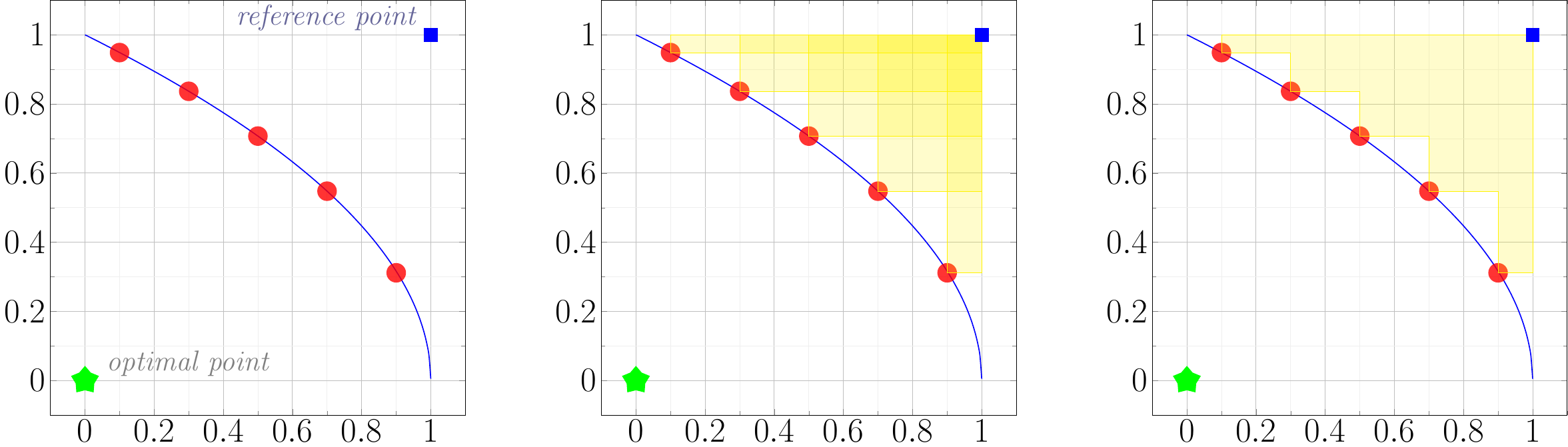}
    \caption{Visual Explanation of Hypervolume. The metric captures the union of axis-aligned rectangles defined by the reference point (star) and the corresponding sample points (red circles). This example showcases loss and the perfect oracle lies in the origin. The point \( (1,1) \) is used for reference.  Hence, higher hypervolume implies that the objective space is better explored/covered.}
    \label{fig:hv}
\end{figure*}

HyperVolume is a metric widely used in multi-objective optimization that captures the quality of exploration. A visual explanation of the metric is given in \Autoref{fig:hv}. \Autoref{tab:hv:small experiments} presents the results of \Autoref{fig:small-experiments} of the main text in a tabular form. We present the best three results per column (higher is better) to succinctly and visually show that all \pmlfull seeds outperform the baselines. 

\newcommand{\goodone}{\cellcolor{green!75}}
\newcommand{\badone}{\cellcolor{red!55}}
\begin{table}[htpb]
    \centering
    \small
    \caption{Tabular complement to \Autoref{fig:small-experiments}. Classification accuracy for both tasks and HyperVolume (HV) metric (higher is better). Three random seeds per method. For baselines, we show the mean accuracy and HV (across seeds). For \pml, we show the results per seed; HV and max accuracies for the subspace yielded by that seed. We use underlined bold, solely bold and solely underlined font for the best, second best and third best results. We observe that the best results are concentrated in the rows concerning the proposed method (\pml). Note that the use of three decimals leads to ties.}
    \label{tab:hv:small experiments}
    \begin{tabular}{lrrrlrrr}
      \toprule
        & \multicolumn{4}{c}{\multimnist} & \multicolumn{3}{c}{\census} \\
        \cmidrule{2-4}
        \cmidrule{6-8}
        &                       Task 1 &                       Task 2 & \multicolumn{2}{l}{HV} &                      Task 1 &                       Task 2 &                           HV \\
      \midrule
      LS             &                        0.955 &                        0.944 &                       0.907 &   &                       0.827 &                        0.785 &                        0.651 \\
      UW             &                        0.957 &                        0.945 &                       0.913 &   &                       0.827 &                        0.785 &                         0.65 \\
      MGDA           &                        0.956 &                        0.943 &                       0.904 &   &                       0.828 &                        0.785 &                        0.651 \\
      DWA            &                        0.955 &                        0.945 &                       0.907 &   &                       0.828 &                        0.785 &                        0.651 \\
      PCGrad         &                        0.955 &                        0.946 &                       0.908 &   &                       0.828 &                        0.785 &                         0.65 \\
      IMTL           &                        0.958 &                        0.944 &                       0.908 &   &                       0.828 &                        0.786 &                        0.651 \\
      Nash-MTL       &                        0.958 &                        0.948 &                       0.913 &   &                       0.827 &                        0.785 &                         0.65 \\
      RLW            &                        0.954 &                        0.941 &                       0.903 &   &                       0.827 &                        0.786 &                        0.651 \\
      Graddrop       &                        0.954 &                        0.942 &                       0.903 &   &           \underline{0.829} &                        0.786 &                        0.652 \\
      Auto-$\lambda$ &                        0.959 &                        0.946 &           \underline{0.918} &   &                       0.827 &                        0.786 &                        0.651 \\
      RotoGrad       &                        0.959 &                        0.945 &                       0.913 &   &                       0.827 &                        0.786 &                        0.651 \\
      \midrule
      PML - 0        & $\underline{\mathbf{0.968}}$ &            \underline{0.951} & $\underline{\mathbf{0.92}}$ &   & $\underline{\mathbf{0.83}}$ & $\underline{\mathbf{0.789}}$ & $\underline{\mathbf{0.655}}$ \\
      PML - 1        &            \underline{0.961} & $\underline{\mathbf{0.953}}$ &                       0.916 &   & $\underline{\mathbf{0.83}}$ & $\underline{\mathbf{0.789}}$ & $\underline{\mathbf{0.655}}$ \\
      PML - 2        &             $\mathbf{0.964}$ & $\underline{\mathbf{0.953}}$ &            $\mathbf{0.919}$ &   &           \underline{0.829} &            \underline{0.788} &            \underline{0.653} \\
      \bottomrule
      \end{tabular}
      
  \end{table}
        \conditionalclearpage
        \subsection{\multimnistThree quantitative results}  
\label{sec:multimnist3:extra experiments}

This section serves as supplementary to \Autoref{sec:beyond} of the main text. \multimnistThree is a synthetic dataset generated by \mnist samples  in a manner similar to the creation of the \multimnist dataset, which is ubiquitous in the \mtl literature. Specifically, each \multimnistThree sample is created with the following procedure. Three randomly sampled digits of size \( 28\times28 \) are placed in the top-left, top-right and bottom middle pixels of a \( 42\times42 \) grid. For the pixels where the initial digits overlap, the maximum value is selected. Finally, the image is resized to \( 28\times28 \) pixels. \Autoref{fig:multimnist3:examples} shows some examples of the dataset, which consists of three digit classification tasks.

\Autoref{tab:multimnist3:baselines} compares the performance of baselines and the proposed method while \Autoref{fig:multimnist3} presents visually the performance achieved on the discovered subspace.

\begin{figure*}[t]
    \centering
    \includegraphics[width=0.95\textwidth]{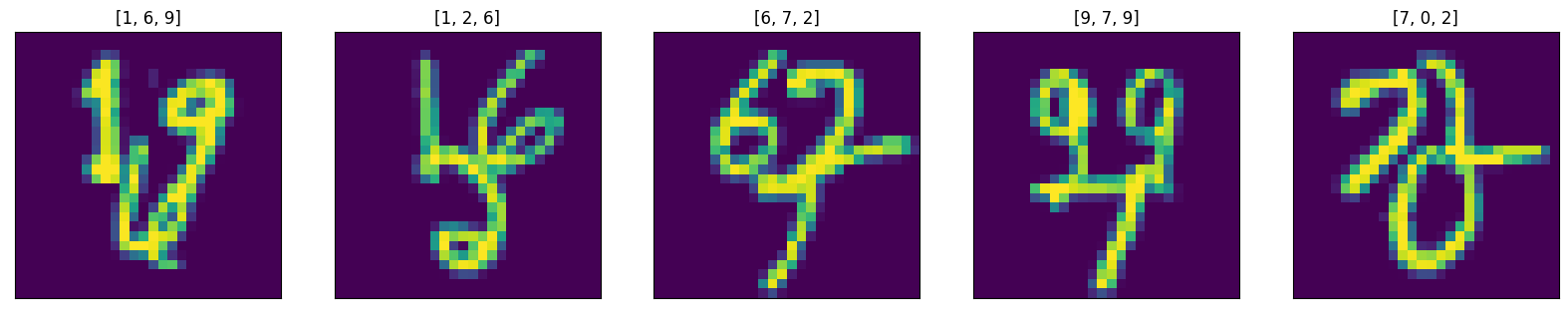}
    \caption{Examples of samples and corresponding labels for the \multimnistThree dataset.}
    \label{fig:multimnist3:examples}
\end{figure*}

\begin{table*}[h]
    \centering
    \caption{\multimnistThree: Mean Accuracy and standard deviation of accuracy  (over 3 random seeds). For the proposed method (\pml), we report the mean and standard deviation of the best performance from the interpolated models in the sampled subspace. No balancing schemes and regularization are applied. Bold is used for the best performing multi-task method.}
    \label{tab:multimnist3:baselines}
    \begin{tabular}{lccc}
        \toprule
        {} &            Task 1 &            Task 2 &            Task 3 \\
        \midrule
        STL      &  96.97 $\pm$ 0.06 &  96.10 $\pm$ 0.17 &  96.40 $\pm$ 0.22 \\
        \midrule
        LS       &  $96.26 \pm 0.20$ &  $95.48 \pm 0.14$ &  $95.87 \pm 0.37$ \\
        UW       &  $96.48 \pm 0.08$ &  $95.42 \pm 0.30$ &  $95.77 \pm 0.06$ \\
        MGDA     &  $96.50 \pm 0.20$ &  $94.80 \pm 0.22$ &  $95.71 \pm 0.08$ \\
        PCGrad   &  $96.45 \pm 0.06$ &  $95.39 \pm 0.15$ &  $95.88 \pm 0.01$ \\
        IMTL     &  $96.58 \pm 0.22$ &  $95.18 \pm 0.12$ &  $96.08 \pm 0.31$ \\
        Graddrop &  $96.25 \pm 0.36$ &  $95.32 \pm 0.24$ &  $95.61 \pm 0.15$ \\
        CAGrad   &  $96.70 \pm 0.13$ &  $95.20 \pm 0.26$ &  $95.66 \pm 0.06$ \\
        RLW      &  $96.06 \pm 0.40$ &  $94.89 \pm 0.18$ &  $95.68 \pm 0.26$ \\
        Nash-MTL &  $\mathbf{96.85 \pm 0.08}$ &  95.25 $\pm$ 0.23 &  96.18 $\pm$ 0.13 \\
        Auto-$\lambda$ &  $96.60 \pm 0.17$ &  $95.16 \pm 0.14$ &  $96.04 \pm 0.18$ \\
        RotoGrad &  $94.80 \pm 0.75$ &  $92.79 \pm 0.87$ &  $94.77 \pm 0.38$ \\
        \midrule
        \pml (ours) & $\mathbf{96.85 \pm 0.43}$ & $\mathbf{95.72 \pm 0.22}$ & $\mathbf{96.27 \pm 0.32}$ \\
        \bottomrule
        \end{tabular}
\end{table*}

\def\separation{10pt}
\newcommand{\plotmultimnistthree}{
\begin{figure*}[t]
    \centering
    \def\qwidth{1}
    \def\bbbeta{0}
    \begin{subfigure}[b]{\qwidth\textwidth}
        \centering
        \includegraphics[width=\textwidth]{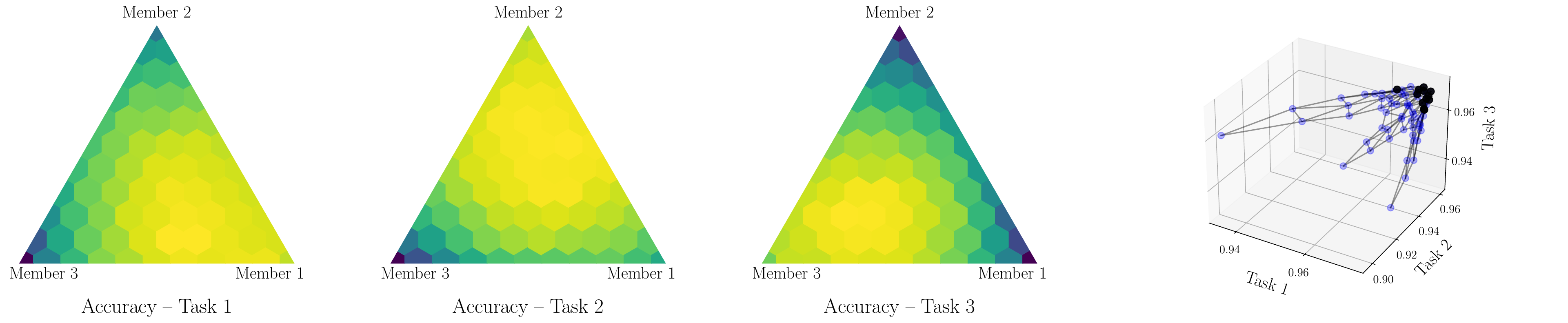}
        \caption{Seed 0}
    \end{subfigure}
    \\[\separation]
    \begin{subfigure}[b]{\qwidth\textwidth}
        \centering
        \includegraphics[width=\textwidth]{media/experiments/multimnistThree/v2_ls-seed=1.pdf}
        \caption{Seed 1}
    \end{subfigure}
    \\[\separation]
    \begin{subfigure}[b]{\qwidth\textwidth}
        \centering
        \includegraphics[width=\textwidth]{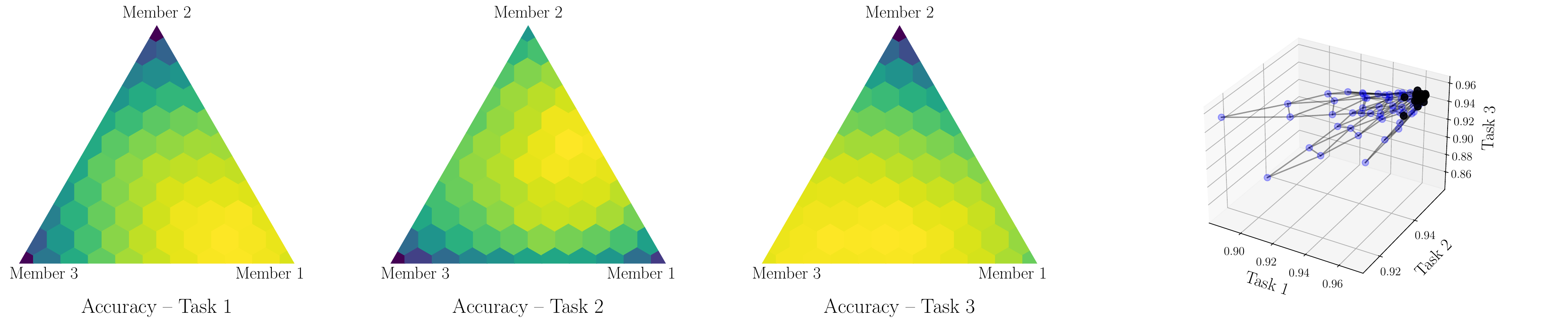}
        \caption{Seed 2}
    \end{subfigure}
    \caption{\multimnistThree results for all three seeds.  
    Each triangle shows the 66 points in the convex hull and color is used for the  performance on the associated task. The 3d plot shows the mapping of the subspace to the multi-objective space. No balancing scheme is used.
    }
    \label{fig:multimnist3}
\end{figure*}
}

\plotmultimnistthree
        \conditionalclearpage
        \subsection{\cityscapes additional results}  
\label{sec:appendix:cityscapes:extra experiments}

The \cityscapes dataset has 2975 training images and no publicly available test set. The validation set is used for test. We refer to the original validation set as test set to avoid confusion.
As far as we understand, prior works in \mtl do not discuss any splitting of the training set to accommodate a validation set. Hence, it is unclear how hyperparameters are set. For this reason, we evaluate on two settings:
\begin{itemize}
    \item Setting I: no validation set. All 2975 images are used for training.
    \item Setting II: Use 500 out of 2975 images for validation. The validation set is used to tune hyperparameters.
\end{itemize} 
The test set is the same in both settings. In the main text, we report the results for Setting II. \autoref{tab:cityscapes-full} presents the results for both settings. For clarity, \pml (ours) uses the same hyperparameters for both settings. While the increase in number of training samples leads to a quantitative boost in performance, the results are qualitatively similar. Specifically, MGDA still performs optimally (out of \shortmtl methods) in Depth Estimation but performs poorly for Segmentation. COSMOS exhibits task bias towards Segmentation. On the other hand, the proposed method produces balanced solutions.

\setlength{\tabcolsep}{5pt}

\begin{table}[t]
\footnotesize
\renewcommand{\arraystretch}{0.95}
\centering
\caption{Test performance on \textit{CityScapes}. See text for description of Settings I and II. 3 random seeds per method. For \pmlfull, we report the mean (across seeds) best results from the final subspace. Methods are divided into single-task, single-solution \abbrevmtl, multi-solution \abbrevmtl and proposed method.} 
\begin{tabular}{@{}lcccccccccccccccccc@{}}
\toprule
{} & \multicolumn{5}{c}{Setting I} & & \multicolumn{5}{c}{Setting II} \\
	\cmidrule(lr){2-6} \cmidrule(lr){8-12}
  & \multicolumn{2}{c}{Segmentation} &                      & \multicolumn{2}{c}{Depth} && \multicolumn{2}{c}{Segmentation} &                      & \multicolumn{2}{c}{Depth} \\
	\cmidrule(lr){8-9} \cmidrule(lr){11-12}
	\cmidrule(lr){2-3} \cmidrule(lr){5-6}
   & \multicolumn{1}{l}{mIoU $\uparrow$} & \multicolumn{1}{l}{Pix Acc $\uparrow$} & \multicolumn{1}{l}{} & \multicolumn{1}{l}{Abs Err $\downarrow$} & \multicolumn{1}{l}{Rel Err $\downarrow$} & \multicolumn{1}{l}{} & \multicolumn{1}{l}{mIoU $\uparrow$} & \multicolumn{1}{l}{Pix Acc $\uparrow$} & \multicolumn{1}{l}{} & \multicolumn{1}{l}{Abs Err $\downarrow$} & \multicolumn{1}{l}{Rel Err $\downarrow$} \\
\midrule
STL                 &      71.79 &        92.60 &&         0.0135 &         32.786 &   &      70.96 &        92.12 &&         0.0141 &         38.644 \\
\midrule
LS                  &      70.94 &        92.29 &&         0.0192 &        117.658 &   &      70.12 &        91.90 &&         0.0192 &        124.061 \\
UW                  &      70.97 &        92.24 &&         0.0188 &        118.168 &   &      70.20 &        91.93 &&         0.0189 &        125.943 \\
MGDA                &      69.23 &        91.77 &&         0.0138 &         51.986 &   &      66.45 &        90.79 &&         0.0141 &         53.138 \\
DWA                 &      70.87 &        92.23 &&         0.0190 &        113.565 &   &      70.10 &        91.89 &&         0.0192 &        127.659 \\
PCGrad              &      71.14 &        92.32 &&         0.0185 &        117.797 &   &      70.02 &        91.84 &&         0.0188 &        126.255 \\
IMTL                &      71.54 &        92.47 &&         0.0151 &         65.058 &   &      70.77 &        92.12 &&         0.0151 &         74.230 \\
Graddrop            &      71.28 &        92.41 &&         0.0182 &        124.645 &   &      70.07 &        91.93 &&         0.0189 &        127.146 \\
CAGrad              &      70.23 &        92.06 &&         0.0173 &        100.162 &   &      69.23 &        91.61 &&         0.0168 &        110.139 \\
RLW                 &      69.94 &        91.94 &&         0.0195 &        119.667 &   &      68.79 &        91.52 &&         0.0213 &        126.942 \\
Nash-MTL            &      72.07 &        92.61 &&         0.0147 &         62.980 &   &      71.13 &        92.23 &&         0.0157 &         78.499 \\
RotoGrad            &      70.41 &        92.03 &&         0.0134 &         48.366 &   &      69.92 &        91.85 &&         0.0193 &        127.281 \\
Auto-$\lambda$      &      71.08 &        92.24 &&         0.0173 &        118.959 &   &      70.47 &        92.01 &&         0.0177 &        116.959 \\
\midrule
COSMOS              &      70.37 &        92.07 &&         0.0317 &        107.575 &   &      69.78 &        91.79 &&         0.0539 &        136.614 \\
\midrule
PaMaL(W=3, $p_0$=7) &     71.13 &        92.31 &&         0.0138 &         50.985 &   &      70.35 &        91.99 &&         0.0141 &         54.520 \\
\bottomrule
\end{tabular}

\label{tab:cityscapes-full}
\end{table}

\fi

\end{document}